\documentclass[10pt,twocolumn,letterpaper]{article}

\usepackage{iccv}
\usepackage{times}
\usepackage{epsfig}
\usepackage{graphicx}
\usepackage{amsmath}
\usepackage{amssymb}

\usepackage{float}
\usepackage{subcaption}
\usepackage{array}
\usepackage{stmaryrd}

\usepackage{multirow}

\usepackage[pagebackref=true,breaklinks=true,letterpaper=true,colorlinks,bookmarks=false]{hyperref}

\newcommand{\myparagraph}[1]{\par\vspace{\baselineskip}\noindent\textbf{#1} }

\newcommand{\real}[1]{\ensuremath{\mathbb{R}^{#1}}}

\newcommand{\Ereal}{\ensuremath{\mathbb{E}_{x \sim p_{data}}}}
\newcommand{\Efake}{\ensuremath{\mathbb{E}_{z \sim p_z}}}
\newcommand{\ErealC}{\ensuremath{\mathbb{E}_{x \sim p_{data}, c \sim p_c}}}
\newcommand{\EfakeC}{\ensuremath{\mathbb{E}_{z \sim p_z, c \sim p_c}}}

\DeclareMathOperator*{\argmin}{arg\,min}

\iccvfinalcopy 

\def\iccvPaperID{****} 

\ificcvfinal\fi

\begin{document}

\title{A Decoupled 3D Facial Shape Model by Adversarial Training}

\author{Victoria Fern\'andez Abrevaya$^1$, Adnane Boukhayma$^2$, Stefanie Wuhrer$^1$, Edmond Boyer$^1$ \\
$^1$ Inria - Univ. Grenoble Alpes - CNRS - LJK, France\\
{\tt\small \{victoria.fernandez-abrevaya, stefanie.wuhrer, edmond.boyer\}@inria.fr}\\
$^2$ University of Oxford, UK\\
{\tt\small adnane.boukhayma@eng.ox.ac.uk}
}

\maketitle
\ificcvfinal\thispagestyle{empty}\fi

\begin{abstract}
Data-driven generative 3D face models are used to compactly encode facial shape data into meaningful parametric representations. A desirable property of these models is their ability to effectively decouple natural sources of variation, in particular identity and expression. While factorized representations have been proposed for that purpose, they are still limited in the variability they can capture and may present modeling artifacts when applied to tasks such as expression transfer.  In this work, we explore a new direction with Generative Adversarial Networks and show that they contribute to better face modeling performances, especially in decoupling natural factors, while also achieving more diverse samples. To train the model we introduce a novel architecture that combines a 3D generator with a 2D discriminator that leverages conventional CNNs, where the two components are bridged by a geometry mapping layer. We further present a training scheme, based on auxiliary classifiers,  to explicitly disentangle identity and expression attributes. Through quantitative and qualitative results on standard face datasets, we illustrate the benefits of our model and demonstrate that it outperforms competing state of the art methods in terms of decoupling and diversity.

\end{abstract}

\section{Introduction}
Generative models of 3D shapes are widely used for their ability to provide compact representations that allow  to synthesize realistic shapes and their variations according to natural factors. This is particularly true with faces whose 3D shape spans a low dimensional space, and for which  generative models often serve as strong priors to solve under-constrained problems such as reconstruction from partial data. Given that the facial shape presents natural factors of variation (\eg identity and expression),	 modeling these in a \emph{decoupled} manner is an important aspect, as it allows to incorporate semantic control when performing inference or synthesis tasks. Having interpretable representations in terms of pre-defined factors of variation opens the door to several applications, such as 3D face animation~\cite{Weise11, Ichim15}, accurate expression transfer~\cite{Kim18, Thies16}, recognition~\cite{Amberg08} and artifical data synthesis~\cite{Shamai19}. 

Since the seminal work of Blanz and Vetter~\cite{Blanz99}, numerous approaches have been proposed to build data-driven generative models of the 3D face. Most commonly, variations among different identities are modeled by linear shape statistics such as PCA~\cite{Blanz99, Booth2016}. When expressions need to be considered the identity and expression subpaces are typically modeled as two independent linear factors which are additively combined~\cite{Amberg08}. In practice this can produce artifacts when transferring expressions among very different facial shapes, an issue  that has to be explicitly accounted for, \eg~\cite{Thies16}. Multilinear models~\cite{Vlasic05, Cao2014, Abrevaya18} present relative improvements by considering a tensor decomposition combining the two spaces, but training requires a complete labeled data tensor which is very hard to get in practice, and transferring expressions by simply switching the latent coefficients can still present artifacts~\cite{Grasshof17}.

With the aim to relax the linear assumption in modeling 3D faces, deep generative models  with autoencoder architectures have recently been proposed. They demonstrate benefits in modeling geometric details~\cite{Bagautdinov18},  non-linear deformations present in facial expressions~\cite{Ranjan18}, and increasing robustness to different types of capture noise~\cite{Abrevaya18}. Yet, none of these approaches decouple the factors of variation with the exception of \cite{Abrevaya18}, where an initialization with fully labeled data is required whose size increases exponentially in the number of considered factors.

In this work we investigate the use of Generative Adversarial Networks (GANs)~\cite{Goodfellow14} for 3D face modeling and provide insights on their ability to learn decoupled representations. 
In particular, our comparisons with recent approaches based on autoencoder architectures~\cite{Abrevaya18,Ranjan18}  demonstrate that our proposed approach can better decouple identity and expression, and exhibit more variability in the generated data.

While current deep learning techniques have shown impressive results in the image domain, extending these to 3D data is not straightforward. 
We propose a novel 3D-2D architecture in which a multilayer perceptron generates a 3D face shape given a latent code, while a regular convolutional network is used as a 2D discriminator. This is allowed by an intermediate \emph{geometry mapping} layer that transforms a 3D surface mesh into a geometry image encoding the mesh vertex locations. To effectively decouple the factors of variation  we build on auxiliary classifiers~\cite{Odena2017} that aim to correctly guess the label associated with each factor, \eg identity and expression, and introduce a loss on the classifier features for unlabeled samples.

To summarize, our contributions are:
\begin{enumerate}
\vspace{-0.5em}
\itemsep0em
\item A generative 3D face model that captures non-linear deformations due to expression, as well as the relationship between identity and expression subspaces.
\item A novel 3D-2D architecture that allows to generate 3D meshes while leveraging the discriminative power of CNNs, by introducing a \emph{geometry mapping layer} that acts as bridge between the two domains.
\item A training scheme that enables to effectively decouple the factors of variation, leading to significant improvements with respect to the state of the art.
\end{enumerate}

\section{Related Work}

Due to the importance of 3D face modeling for numerous applications, many works have been proposed to learn generative models. We focus here on data-driven approaches, often called \emph{3D Morphable Models} (3DMM) in the literature. Blanz and Vetter~\cite{Blanz99} use principal component analysis (PCA) to learn the distribution of the facial shape and appearance across different identities scanned in a neutral expression. To handle other expressions, subsequent works model them by either adding linear factors~\cite{Amberg08} or by extending PCA to a multilinear model~\cite{Vlasic05}. Thanks to their simple structure these models are still heavily used, and have recently been extended by training from large datasets~\cite{Booth2016}, modeling geometric details~\cite{Neumann2013,Brunton2014a,Cao2015}, and including other variations such as skeletal rotations~\cite{Li2017}. 

\myparagraph{Autoencoders for 3D Faces}
Recent works leverage deep learning methods to overcome the limitations of (multi-)linear models. Ranjan~\etal~\cite{Ranjan18} proposed an autoencoder architecture that learns a single global model of the 3D face, and as such the different factors cannot be decoupled directly. However, an extension called DeepFLAME is proposed that combines a linear model of identity~\cite{Li2017} with the autoencoder trained on expression displacements. While expressions are modeled non-linearly, the relationship between identity and expression is not addressed explicitly. Fern\'{a}ndez Abrevaya~\etal~\cite{Abrevaya18} developed a multilinear autoencoder (MAE) in which the decoder is a multilinear tensor structure. While the relationship between the two spaces is accounted for, transferring expressions still presents artifacts. Furthermore, to achieve convergence the tensor needs to be initialized properly, which implies that the size of labeled training data needed for initialization increases exponentially in the number of factors considered. We compare our proposed approach to DeepFLAME and MAE, as they achieve state-of-the-art results on decoupling identity and expression variations. 

Bagautdinov~\etal~\cite{Bagautdinov18} propose a multiscale model of 3D faces at different levels of geometric detail. Two recent works~\cite{Tran18,Tewari18} use autoencoders to learn a global or corrective morphable model of 3D faces and their appearance based on 2D training data. However, none of these methods allow to disentangle  factors of variation in the latent space. Unlike the aforementioned works, we investigate the use of GANs to learn a decoupled model of 3D faces.

\myparagraph{GANs for 3D faces} Some recent works have proposed to combine a 3DMM with an \emph{appearance} model obtained by adversarial learning. Slossberg~\etal\cite{Slossberg18} train a GAN on aligned facial textures and combine this with a linear 3DMM to generate realistic synthetic data. Gecer~\etal\cite{Gecer19} train a similar model and show that GANs can be used as a texture prior for accurate fitting to 2D images. Deng~\etal~\cite{Deng18} fit a 3DMM to images and use a GAN to complete the missing parts of the resulting UV map.  All of these methods rely on linear 3DMMs, and hence to shape spaces limited in expressiveness. While the focus is on improving the appearance, we follow a different objective with a generative shape model that decouples identities and expressions. 

To the best of our knowledge, the only work that learns 3D facial \emph{shape} variations using a GAN is~\cite{Shamai19}, which is an extension of~\cite{Slossberg18}. The authors propose to learn identity variations by training a GAN on geometry images, but unlike our work they do not model the non-linear variations due to expression nor the correlation between identity and expression, since the main focus is on the appearance.

Two other methods learn to enhance an input 3D face geometry with photometric information using a GAN. Given a texture map and a coarse mesh, Huynh~\etal~\cite{Huynh18} augment the latter with fine scale details, and given an input image and a base mesh, Yamaguchi~\etal~\cite{Yamaguchi18} infer detailed geometry and high quality reflectance. Both works require the conditioning of an input, and unlike us they do not build a generative 3D face model.

\section{Background}

\begin{figure*}[t]
\begin{center}
\includegraphics[width=.9\textwidth]{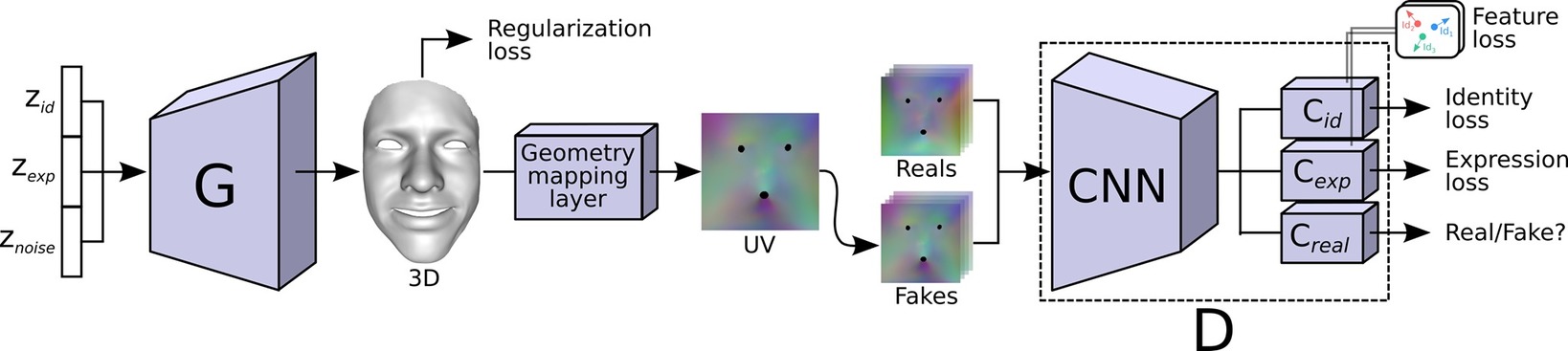}
\caption{Our proposed architecture.  A MLP generates the 3D coordinates of the mesh, while the discrimination occurs in 2D space thanks to the \emph{geometry mapping} layer. Identity and expression codes $z_{id}, z_{exp}$ are used to control the generator, and classification losses are added to decouple between the two. A feature loss is introduced to ensure consistency over features with fixed identities or expressions.}
\label{fig:method}
\end{center}
\end{figure*}

Generative Adversarial Networks~\cite{Goodfellow14} are based on a minimax game, in which a discriminator $D$ and a generator $G$ are optimized for competing goals. The discriminator is tasked with learning the difference between real and fake samples, while the generator is trained to maximize the mistakes of the discriminator. At convergence, $G$ approximates the real data distribution. Training involves the optimization of the following:
\begin{eqnarray}
\displaystyle
	\min_{G} \max_{D} \mathcal{L}_{GAN} & = & \Ereal [\log D(x)] \nonumber \\
	& + &	\Efake [\log (1 - D(G(z)))],
	\label{eq:gan_loss1}
\end{eqnarray}
where $p_{data}$ denotes the distribution of the training set, and $p_z$ denotes the prior distribution for G, typically $\mathcal{N}(0, I)$. 

GANs have been shown to be very challenging to train with the original formulation and prone to low diversity in the generated samples. To address this, Arjovsky~\etal~\cite{Arjovsky17} propose to minimize instead an approximation of the Earth Mover's distance between generated and real data distributions, which is the strategy we adopt in this work:
\begin{eqnarray}
\displaystyle
	\mathcal{L}_{GAN} & = & \Ereal [D(x)] -\Efake [D(G(z))].
	\label{eq:gan_loss}
\end{eqnarray}
In particular we use the extension in~\cite{Gulrajani17} which uses a gradient penalty in order to enforce that $D$ is 1-Lipschitz.

When labels are available, using them has proven to be beneficial for GAN performance. Odena~\etal~\cite{Odena2017} proposed Auxiliary Classifier GANs (AC-GAN), in which $D$ is augmented so that it  outputs the probability of an image belonging to a pre-defined class label $c \sim p_c$. In this case, the loss function for G and D is extended with:
\begin{equation}
\mathcal{L}_C^{real} = \ErealC [ \log P(C = c | x) ],
\label{eq:real_label_loss}
\end{equation}
\begin{equation}
\label{eq:C_fake}
\mathcal{L}_C^{fake} = \EfakeC [ \log P(C = c | G(z, c)) ].
\end{equation}

\medskip
In order to evaluate if a model is correctly decoupling, we need to be able to distinguish whether two identites or expressions sharing the same latent code are perceptually similar. Thus, our work builds on the idea of auxiliary classifiers in order to learn a decoupling of the shape variations into factors, as will be explained in the next section.

\section{Method}

We consider as input a dataset of registered and rigidly aligned 3D facial meshes, where each mesh is defined by $( \mathcal{V}, \mathcal{F} )$, the set of 3D vertices $\mathcal{V} \in \real{3 \times n_v}$ and the set of triangular faces $\mathcal{F} \in \mathbb{N}^{3 \times n_f}$ that connect the vertices.
Our goal is to build an expressive model that can decouple the representation based on known factors of variation. 
In contrast to classical approaches in which a reconstruction error is optimized, we rely instead on the adversarial loss enabled by a convolutional discriminator. To this end, we introduce an architecture in which a \emph{geometry mapping layer} serves as bridge between the generated 3D mesh and the 2D domain, for which convolutional layers can be applied (Section~\ref{sec:geom_layer}). To learn a decoupled parameterization, we build on the idea of Auxiliary Classifiers and introduce a feature loss to further improve the results (Section~\ref{sec:method_train}). We will consider here a model that decouples between identity and expression, however the principle can be easily extended to more factors.

\subsection{Geometry Mapping Layer}
\label{sec:geom_layer}

\begin{figure}[b]
\centering
 
\begin{subfigure}[b]{.15\textwidth}
	\centering
	\includegraphics[width=0.7\textwidth]{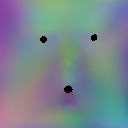} 
\caption{Geometry image}
\label{fig:imrec_a}
\end{subfigure}
\begin{subfigure}[b]{.3\textwidth}
	\centering
	\includegraphics[width=.9\textwidth]{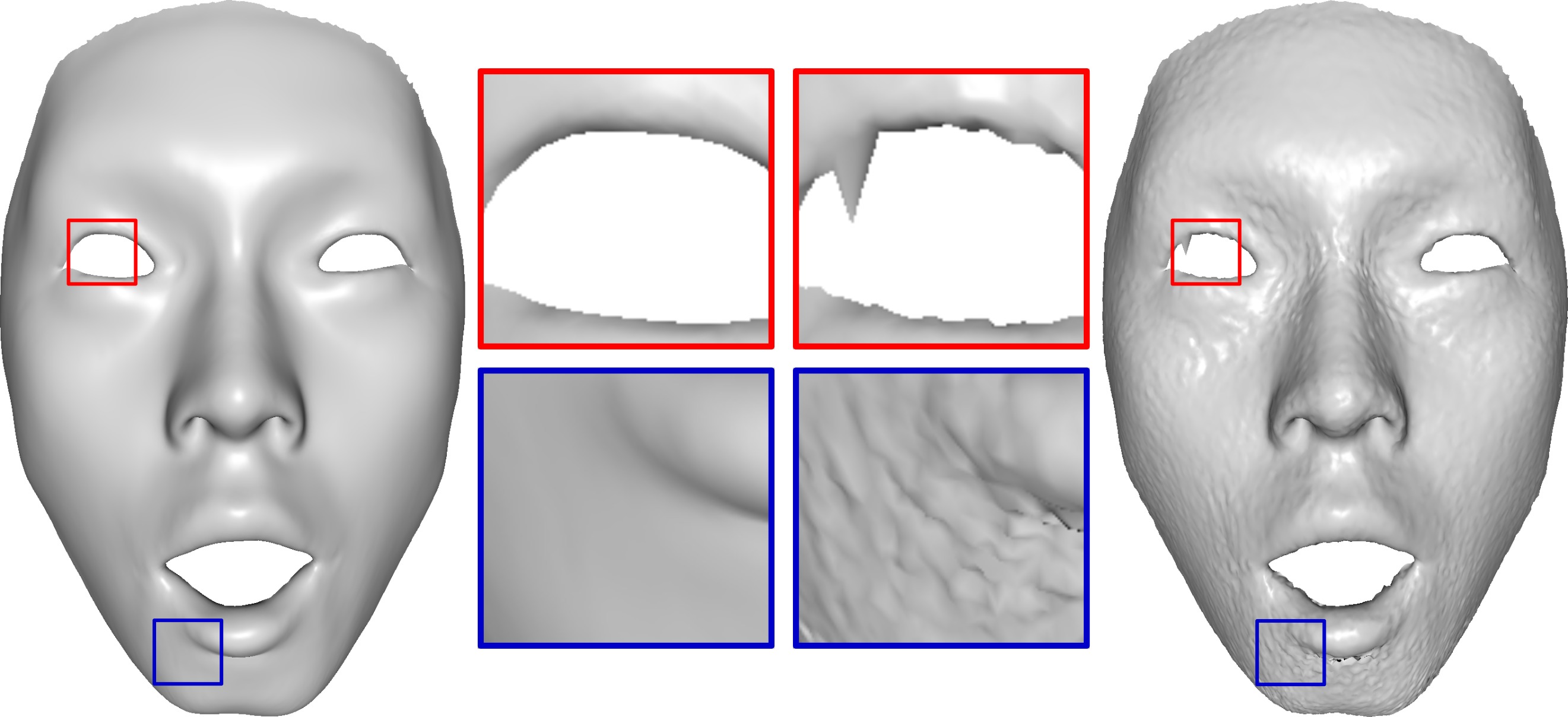} 
\caption{Original and reconstructed meshes}
\label{fig:imrec_b}
\end{subfigure}
\caption{While a GAN could be used to generate geometry images, recovering the mesh from them is prone to artifacts, e.g. erroneous boundary interpolations (red) and precision loss (blue) in \ref{fig:imrec_b}. In this work we generate instead the 3D mesh, while geometry images are used only for discrimination.}
\label{fig:imrec}
\end{figure}

While deep learning can be efficiently used on regularly sampled signals, such as 2D pixel grids, applying it to 3D surfaces is more challenging due to their irregular structure. In this work we propose to generate the 3D coordinates of the mesh using a multilayer perceptron, while the discriminative aspects are handled in the 2D image domain. This allows to benefit from efficient and well established architectures that have been proven to behave adequately under adversarial training, while still generating the 3D shape in its natural domain. 

In particular, a 2D representation of a mesh can be achieved through a UV parameterization $\phi : \mathcal{V} \rightarrow D$ that associates each vertex $v_k \in \mathcal{V}$ with a coordinate $(u,v)_k$ in the unit square domain $D$. Continuous images can be obtained by interpolating the $(x,y,z)$ vertex values according to the 2D barycentric coordinates, and storing them in the image channels. Borrowing the term from~\cite{Gu02}, we call this a \emph{geometry image} (see Figure~\ref{fig:imrec_a}). 

Note that although our method could  generate geometry images instead of 3D meshes, this would introduce an unnecessary additional reconstruction step that is likely to  cause information loss and artifacts in the final meshes, as illustrated in Figure~\ref{fig:imrec_b}. This is due to the fact that a single planar unfolding of a mesh may create distortions such as triangle flipping~\cite{Sheffer06}, and a many-to-one mapping may be obtained even with a bijective parameterization due to the finite size of images. In addition, as elaborated in \cite{Gu02}, unless border vertices are preassigned to distinct pixels which can be challenging for large meshes, sampling these locations results in erroneous interpolations. Generating 3D point coordinates instead allows to avoid reconstruction artifacts, and to apply common mesh regularization techniques that simplify and improve the learning process. We use geometry images only as the representation for the discriminative component that evaluates the 3D generator through CNNs. 

\medskip
The mapping layer operates as follows. Given a mesh made of vertices $\mathcal{V} = \{v_k / k = 1 .. n_v  \}$, a target image size $n \times n$, and a pre-computed UV parameterization $\phi$, we build two images $I^{\mathcal{U}}$, $I^{\mathcal{V}}$ of dimension $n \times n$, and three images $I^{v_1}$, $I^{v_2}$ and $I^{v_3}$ of dimension $n \times n \times 3$ each. For each pixel $(i,j)$, we consider the $\phi$-projected mesh triangle $(\hat{v}_1,\hat{v}_2,\hat{v}_3)$ containing it. The barycentric abscissa and ordinate of pixel $(i,j)$ in triangle $(\hat{v}_1,\hat{v}_2,\hat{v}_3)$ are then stored in images $I^{\mathcal{U}}$ and $I^{\mathcal{V}}$ respectively, and the original face vertex coordinates $v_1$, $v_2$ and $v_3$ are stored in images $I^{v_1}$, $I^{v_2}$ and $I^{v_3}$. The mapping layer computes the output geometry image $\mathcal{I}$ as:
\begin{equation}
\mathcal{I} = I^{\mathcal{U}} * I^{v_1} + I^{\mathcal{V}} * I^{v_2} + (\mathbf{1} -  I^{\mathcal{U}} -  I^{\mathcal{V}}) * I^{v_3},
\end{equation}
where $*$ denotes element-wise multiplication and $\mathbf{1} \in \real{n \times n}$ is the matrix of ones. Since this layer simply performs indexing and linear combinations on the elements of $\mathcal{V}$ using the predefined parameters in $I^\mathcal{U}$ and $I^\mathcal{V}$, all operations are differentiable and the gradients can be back-propagated from the discriminated image to the generated mesh.

\subsection{Architecture}
\label{sec:architecture}
\smallskip
Figure~\ref{fig:method} depicts our proposed architecture. The generator consists of two fully connected layers that map the latent code $z$ to a vector of size $3n_v$ containing the stacked 3D coordinates of displacements from a reference face mesh. The output vertex positions are passed through the mapping layer to generate a geometry image of size $n \times n$, which is then processed by the discriminator in order to classify whether the generated mesh is real or fake. We also consider auxiliary classifiers for the discriminator, denoted as $C_{id}$ and $C_{exp}$. The design of D shows two main differences with respect to the original AC-GAN. First, instead of classifying only one type of labels, we use here classifiers for both identity and expression. This favors decoupling, since the classification of one factor is independent of the choice of the labels for the other factors. Second, we provide distinct convolutional layers for the real/fake, identity and expression blocks. This is motivated by the observation that the features required to classify identities and expressions are not necessarily the same.

\subsection{Decoupled Model Learning}
\label{sec:method_train}

We rely on the discriminator not only to generate realistic faces, but also to decouple the factors of variation. For this, we optimize D such that it maximizes
\begin{equation}
	\mathcal{L}_{D} = \mathcal{L}_{GAN} + \lambda_C (\mathcal{L}_{ID} + \mathcal{L}_{EXP}).
	\label{eq:discriminator}
\end{equation}

Here, $\mathcal{L}_{GAN}$ denotes the standard adversarial loss (see Equation~\ref{eq:gan_loss}), and $\mathcal{L}_{ID}, \mathcal{L}_{EXP}$  the classification losses measured against the labels provided with the dataset and weighted by scalar $\lambda_C$. These losses are defined similarly to Equation~\ref{eq:real_label_loss} as: 
\vspace{-0.5em}
\begin{eqnarray}
\mathcal{L}_{ID}  &=& \mathbb{E}_{x \sim p_{data}, c \sim p_c^{id}} [ \log P(C = c | x) ],\nonumber\\
\mathcal{L}_{EXP} &=& \mathbb{E}_{x \sim p_{data}, c \sim p_c^{exp}} [ \log P(C = c | x) ],
\end{eqnarray}
where $p_c^{id}$ and $p_c^{exp}$ denote the distribution of identity and expression labels, respectively. We ignore the sample contribution in the classification loss if it is not labeled.

\smallskip
The generator $G$ takes as input a random vector $z = \{z_{id}, z_{exp}, z_{noise} \}$, which is the concatenation of the identity code $z_{id} \sim p_{id}$, the expression code $z_{exp} \sim p_{exp}$ and a random noise $z_{noise} \sim p_{noise}$. It produces the location of $n_v$ displacement vectors from a reference mesh, and is trained by minimizing:
\begin{equation}
\begin{split}
\mathcal{L}_G = \lambda_1 \mathcal{L}_{GAN} - \lambda_2 \left(\mathcal{L}_{CL}^{id} + \mathcal{L}_{CL}^{exp}\right) \\
+ \lambda_3 \left(\mathcal{L}_{FEAT}^{id} + \mathcal{L}_{FEAT}^{exp}\right) + \lambda_4 \mathcal{L}_{reg},
\end{split}
\label{eq:generator}
\end{equation}
where $\mathcal{L}_{GAN}$ is the standard GAN loss (Equation~\ref{eq:gan_loss}); $\mathcal{L}_{CL}^{id}$ and $\mathcal{L}_{CL}^{exp}$ are classification losses; $\mathcal{L}_{FEAT}^{id}$ and $\mathcal{L}_{FEAT}^{exp}$ are feature losses that aim to further increase the decoupling of the factors; $\mathcal{L}_{reg}$ is a regularizer; and $\lambda_1, \lambda_2, \lambda_3, \lambda_4$ are weights for the different loss terms.  We explain each of these in the following.

\myparagraph{Classification Loss} In addition to the adversarial loss, the generator is  trained to  classify its samples with  the correct labels by maximizing:
\begin{eqnarray}
\mathcal{L}_{CL}^{id}  &=& \mathbb{E}_{z \sim p_z, c \sim p_c^{id}} [ \log P(C = c | G(z)) ]
\nonumber\\
\mathcal{L}_{CL}^{exp}  &=& \mathbb{E}_{z \sim p_z, c \sim p_c^{exp}} [ \log P(C = c | G(z)) ].
\end{eqnarray}
In order to generate data belonging to a specific class, we sample one identity/expression code $z_{id}, z_{expr}$ for each label and fix it throughout the training; this becomes the input for $G$ each time the classification loss must be evaluated. We denote the set of fixed codes for identity and expression as $\mathcal{T}^{id}$ and $\mathcal{T}^{exp}$ respectively.

\myparagraph{Feature Loss} 
The classification loss is limited to codes in $\mathcal{T}^{id} / \mathcal{T}^{exp}$, which have  associated labels. We found that better decoupling results can be obtained if we include a loss on the classifier features. We measure this by generating samples in pairs which share the same identity or expression vector, and measuring the error as: 
\begin{equation}
\label{eq:feat_loss_id}
\mathcal{L}_{FEAT}^{id} = \frac{2}{N} \sum_{z_{id}} \left( 1 - cos(\mathbf{f}_{1,z_{id}}, \mathbf{f}_{2,z_{id}})  \right),
\end{equation}
\begin{equation}
\label{eq:feat_loss_exp}
\mathcal{L}_{FEAT}^{exp} = \frac{2}{N} \sum_{z_{exp}} \left( 1 - cos(\mathbf{f}_{1,z_{exp}}, \mathbf{f}_{2,z_{exp}})  \right).
\end{equation}
Here, $N$ is the batch size, and $\mathbf{f}_{i,z_{id}} = \mathbf{f} \left( G(z_{id}, z_{exp,i}, z_{noise,i} )\right)$ are feature vectors obtained by inputting the sample $G(z_{id}, z_{exp,i}, z_{noise,i} )$ through the classifier $C_{id}$ and extracting the features from the second to last layer. That is,  given two inputs which were generated with the same identity vector, $\mathcal{L}_{FEAT}^{id}$ enforces that their feature vectors in the identity classifier are also aligned. The definition is analogous for $\mathbf{f}_{i,z_{exp}}$ with $C_{exp}$.

To enable training with both classification and feature loss, for each batch iteration we alternate between the sampling of labeled identity codes $z_{id} \in \mathcal{T}^{id}$ with unlabeled expression codes $z_{exp} \sim p_{exp}$, and the sampling of unlabeled identity codes $z_{id} \sim p_{id}$ with labeled expression codes $z_{exp} \in \mathcal{T}^{exp}$. The  classification is evaluated for the labeled factor only, while the feature loss is used for unlabeled codes, and the alternation allows to better cover the identity and expression sub-spaces during training. 

\myparagraph{Regularization}
Generating a 3D mesh allows us to reason explicitly at the surface level and define high order loss functions using the mesh connectivity. In particular, we enforce spatial consistency over the generated faces by minimizing the following term on the output displacements $\mathbf{v} = G(z)$:
\vspace{-0.5em}
\begin{equation}
\label{eq:reg}
\mathcal{L}_{reg} = || L\mathbf{v} ||_2^2,
\end{equation}

where $L$ is the cotangent discretization of the Laplace-Beltrami operator. 
\section{Results}

We provide in this section results obtained with the proposed framework, which demonstrate its benefits particularly in decoupling. We first clarify our set-up with implementation details in Section~\ref{sec:implementation} and the datasets used in~\ref{sec:datasets}. We explain in Section~\ref{sec:metrics} the proposed metrics for the evaluation of a 3D face model, and introduce a new measure for analyzing the diversity of the generated samples. In Section~\ref{sec:ablation} we perform ablation studies to verify that all the components are necessary to effectively train an expressive model. Finally, in Section~\ref{sec:comparison} we compare our results to state-of-the-art 3D face models that can decouple the latent space, and show that our approach outperforms with respect to decoupling and diversity. Additional  results can be found in the supplemental material.

\subsection{Implementation Details}
\label{sec:implementation}

We set the weights to $\lambda_C=0.1$ (Equation~\ref{eq:discriminator}), $\lambda_1 = \lambda_2 = 1$, $\lambda_3 = 0.5$ and $\lambda_4 = 100$ (Equation~\ref{eq:generator}). The classification losses are further weighted to account for unbalanced labels~\cite{King01}. For the generator, we use two fully connected layers with an intermediate representation of size $512$ and ReLU non-linearity. For the discriminator we use a variant of DC-GAN \cite{Radford16}, with the first two convolutional blocks shared between $C_{real}$, $C_{id}$ and $C_{expr}$, while the remaining are duplicated for each module (more details can be found in supplemental). The models were trained for $200$ epochs using ADAM optimizer~\cite{Kingma14} with $\beta_1 = 0.9$ and $\beta_2 = 0.999$, a learning rate of $0.0002$ and a batch size of $64$. During training we add instance noise~\cite{Sonderby17} with $\sigma=0.1$ to the input of D. The discriminator is trained for $3$ iterations each time we train the generator. The models take around $2$ hours to train on a NVidia GeForce GTX 1080 GPU. 

The template mesh contains $22129$ vertices. We pre-compute the UV map $\phi$ using harmonic parameterization~\cite{Eck95}, setting the outer boundary face vertices to a unit square to ensure full usage of the image domain. We generate geometry images of size $64 \times 64$; we experimented with other image sizes but the best decoupling results were obtained with this resolution. The dimensions for $(z_{id},z_{exp},z_{noise})$ are set to $(65,15,5)$ to facilitate comparison with \cite{Abrevaya18}, and the feature vectors used in Equations~\ref{eq:feat_loss_id} and~\ref{eq:feat_loss_exp} are of size $2048$.

\subsection{Datasets}
\label{sec:datasets}
All models were trained using a combination of four publicly available 3D face datasets. In particular, we use two datasets containing static 3D scans of multiple subjects: BU-3DFE \cite{Yin06} and Bosphorus \cite{Savran08}, and combine these with two datasets of 3D motion sequences of multiple subjects: BP4D-Spontaneous \cite{Zhang14} and BU-4DFE \cite{Yin08}. The static datasets provide variability of identities, while the motion datasets provide variability of expressions and a larger number of training samples. We registered BU-3DFE and Bosphorus with a template fitting approach~\cite{Salazar14}, and the motion datasets with a spatiotemporal approach~\cite{Abrevaya18b}.

The final dataset contains $30559$ registered 3D faces and was obtained by subsampling the motion sequences. We provide identity labels for all meshes, while the expression labels are limited to the seven basic emotional expressions, which appear in both static datasets. For BU-4DFE, expression labels are assigned to three frames per sequence: the neutral expression to the first and last frame, and the labeled expression of the sequence to the peak frame. For BP4D, one neutral frame is manually labeled per subject (this is a requirement for comparison to~\cite{Ranjan18}). Overall, due to the use of motion data, only $7\%$ of it is assigned expression labels.

\subsection{Evaluation Metrics}
\label{sec:metrics}
We evaluate the models in terms of \emph{diversity} of the generated samples, \emph{decoupling} of identity and expression spaces, and \emph{specificity} to the 3D facial shape. We believe it is necessary to simultaneously consider all the metrics, as they provide complementary information on the model. For instance, a good decoupling value can be obtained when the diversity is poor, since small variations facilitate the classification of samples as ``same''. Conversely,  a large diversity value can be obtained when decoupling is poor, since the identities/expressions sharing the same code can yield very different shapes. We detail these in the following.

\myparagraph{Diversity}  We consider it important to measure the diversity of the 3D face shapes generated by a model, particularly with GANs that are known to be prone to mode collapse. To the best of our knowledge, this has not yet been considered in the context of 3D face models and we propose therefore to evaluate as follows. We sample $p$ pairs of randomly generated meshes and compute the mean vertex distance among the pairs; diversity is then defined as the mean of the distances over the $p$ pairs. We expect here to see higher values for more diverse models. We evaluate on three sets of sampled pairs: (1) among pairs chosen randomly (\emph{global diversity}), (2) among pairs that share the same identity code (\emph{identity diversity}) and (3) among pairs that share the same expression code (\emph{expression diversity}). For all cases we evaluate on $10000$ pairs. For comparison, the training set is also evaluated on these three metrics by leveraging the labels. 

\myparagraph{Decoupling} To evaluate decoupling in both identity and expression spaces we follow the protocol proposed in \cite{Donahue18}. In particular, we first train two networks, one for identity and one for expression, that transform an image representation of the mesh to an $n$-dimensional vector using triplet loss~\cite{Schroff15}, where $n=128$ in our experiments. The trained networks allow to measure whether two meshes share the same identity or expression by checking whether the distance between their embeddings is below a threshold $\tau$. 

To measure identity decoupling, we generate $n$ random faces $\mathbf{x}_i = G(z_{id}^i, z_{exp}^i, z_{noise}^i)$, and for each random face we fix the identity code and sample $m$ faces $\mathcal{Y}(\mathbf{x}_i) = \{ G(z_{id}^i, z_{exp}^j, z_{noise}^j), j = 1..m \}$. We then use the embedding networks to evaluate whether the original faces $\mathbf{x}_i$ and their corresponding samples in $\mathcal{Y}(\mathbf{x}_i)$ correspond to the same identity, and report the percentage of times the pairs were classified as ``same''. We proceed analogously for expression decoupling. We set $n=100$, $m=100$, $\tau=0.14$ for identity and $\tau=0.226$ for expression; more implementation details are given in the supplemental material.

\myparagraph{Specificity} Specificity is a metric commonly used for the evaluation of statistical shape models~\cite{Davies08} and whose goal is to quantify whether all the generated samples belong to the original shape class, faces in our case. For this, $n$ samples are randomly drawn from the model and for each the mean vertex distance to each member of the training set is measured, keeping the minimum value. The metric then reports the mean of the $n$ values. We use here $n=1000$.

\subsection{Ablation Tests}
\label{sec:ablation}

\begin{figure*}[t!]
\centering

\begin{tabular}{c c c}
	\includegraphics[height=.075\textheight]{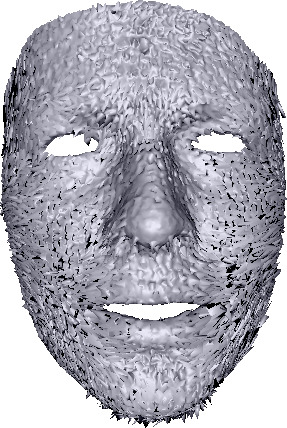}
	\includegraphics[height=.075\textheight]{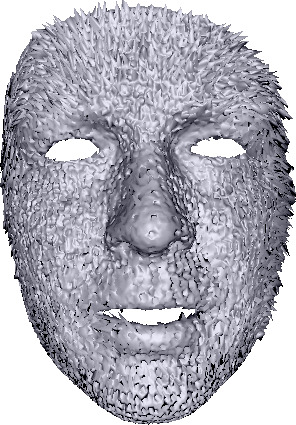}
	\includegraphics[height=.075\textheight]{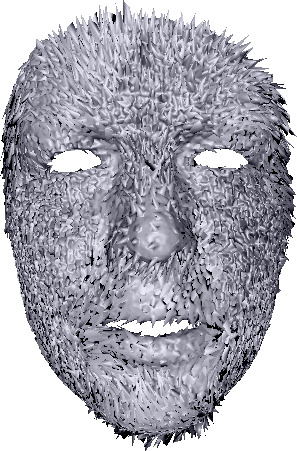}
	\includegraphics[height=.075\textheight]{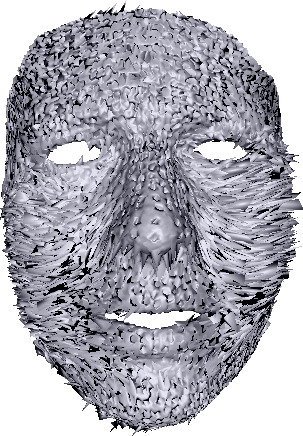} &
	\includegraphics[height=.075\textheight]{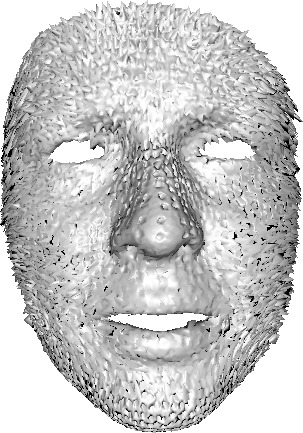}
	\includegraphics[height=.075\textheight]{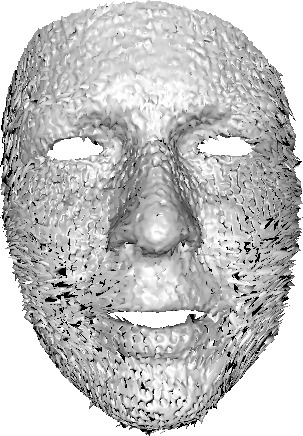}
	\includegraphics[height=.075\textheight]{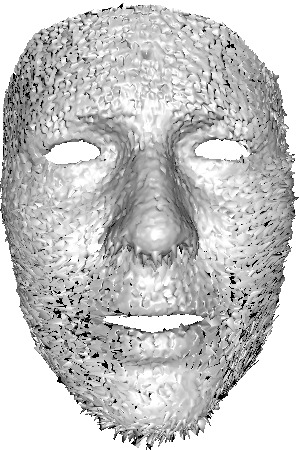}
	\includegraphics[height=.075\textheight]{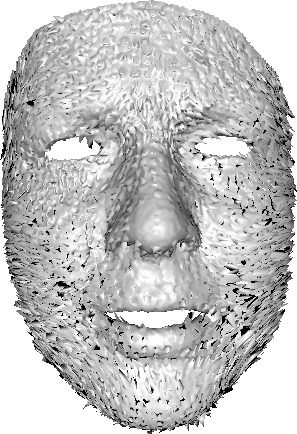} &
	\includegraphics[height=.075\textheight]{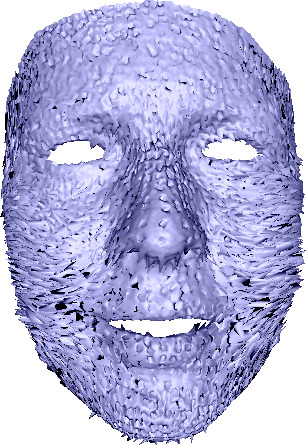}
	\includegraphics[height=.075\textheight]{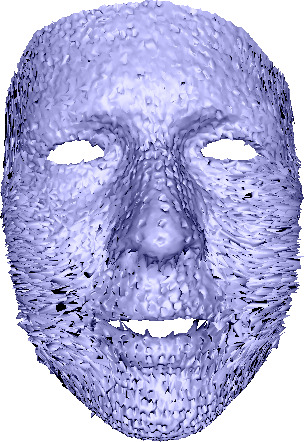}
	\includegraphics[height=.075\textheight]{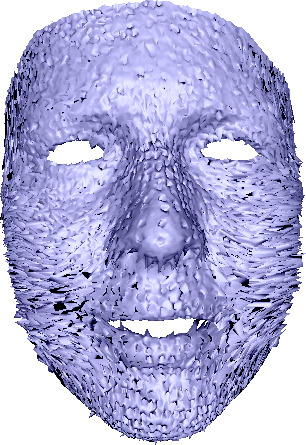}
	\includegraphics[height=.075\textheight]{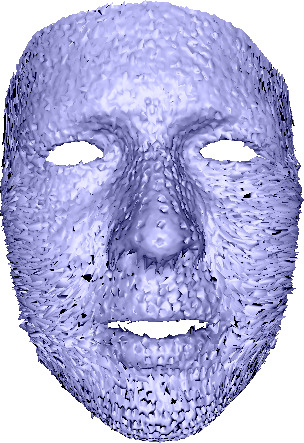} \\
&
\begin{subfigure}[b]{0.3\textwidth}
\caption{Without mesh regularization}
\label{fig:ablation-no-reg}
\end{subfigure}\vspace{-5pt}
&\\
	\includegraphics[height=.075\textheight]{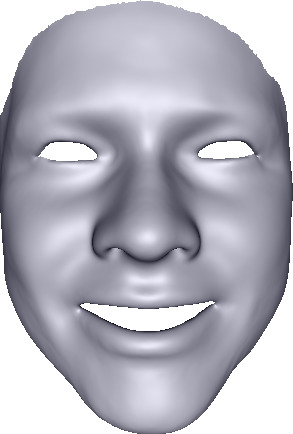}
	\includegraphics[height=.075\textheight]{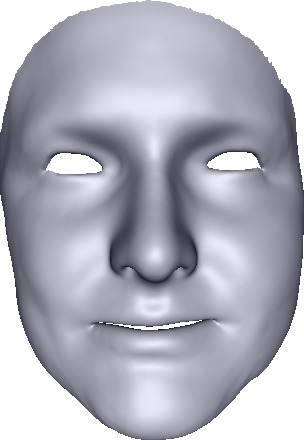}
	\includegraphics[height=.075\textheight]{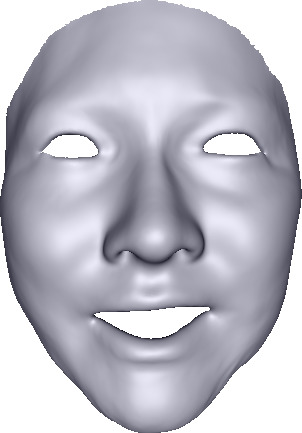}
	\includegraphics[height=.075\textheight]{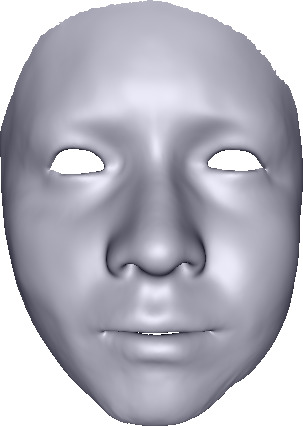} &
	\includegraphics[height=.075\textheight]{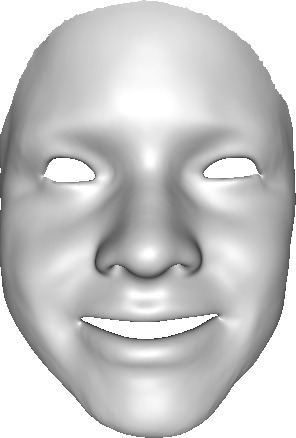}
	\includegraphics[height=.075\textheight]{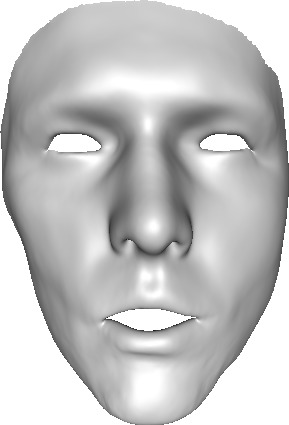}
	\includegraphics[height=.075\textheight]{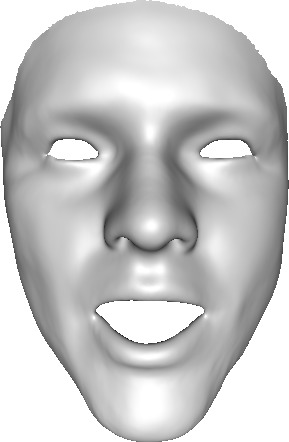}
	\includegraphics[height=.075\textheight]{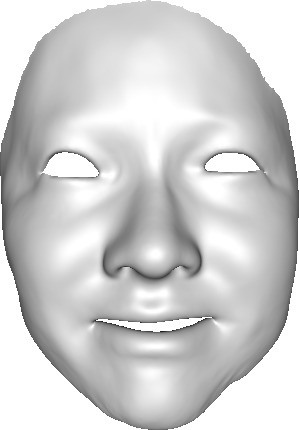} &
	\includegraphics[height=.075\textheight]{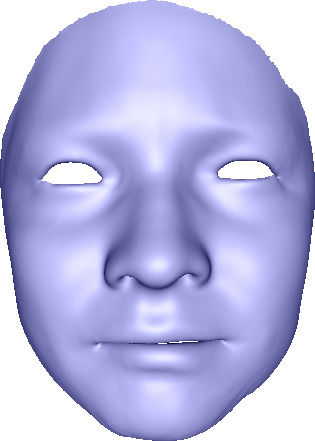}
	\includegraphics[height=.075\textheight]{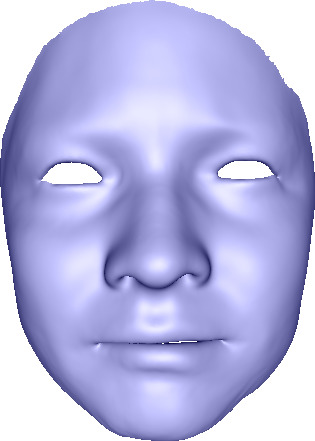}
	\includegraphics[height=.075\textheight]{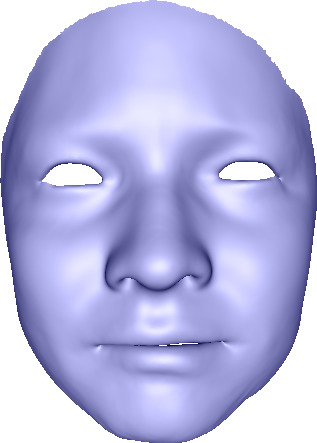}
	\includegraphics[height=.075\textheight]{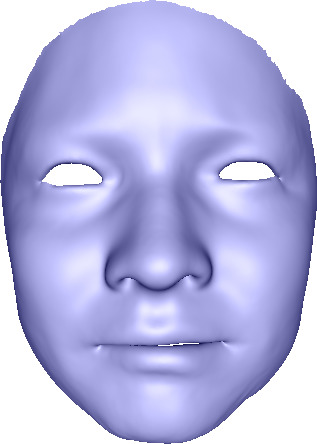} \\

&
\begin{subfigure}[b]{0.3\textwidth}
\caption{Without expression classification}
\label{fig:ablation-no-expr}
\end{subfigure}\vspace{-5pt}
&\\
	\includegraphics[height=.075\textheight]{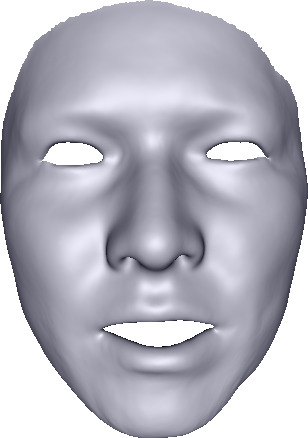}
	\includegraphics[height=.075\textheight]{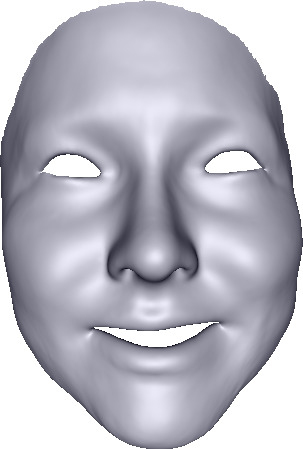}
	\includegraphics[height=.075\textheight]{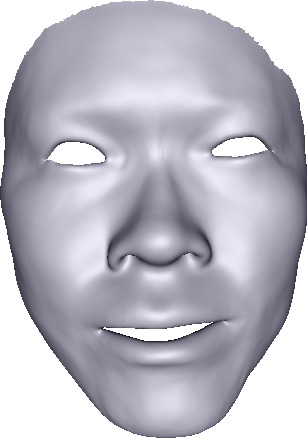}
	\includegraphics[height=.075\textheight]{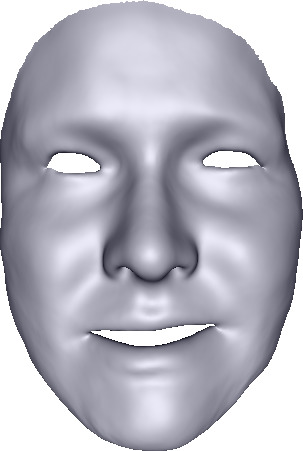} &
	\includegraphics[height=.075\textheight]{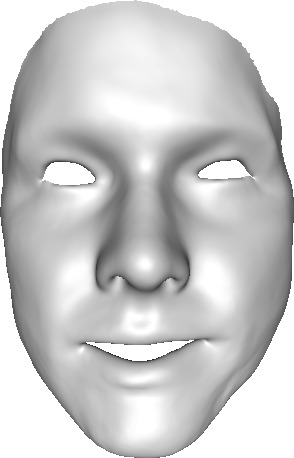}
	\includegraphics[height=.075\textheight]{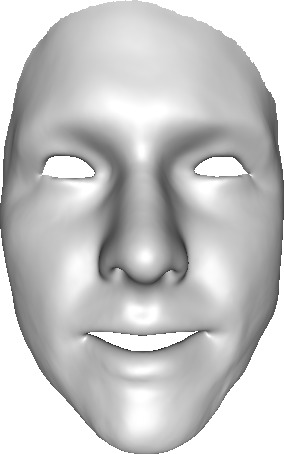}
	\includegraphics[height=.075\textheight]{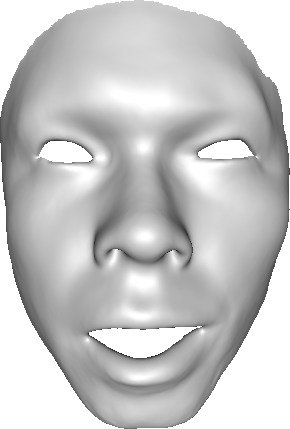}
	\includegraphics[height=.075\textheight]{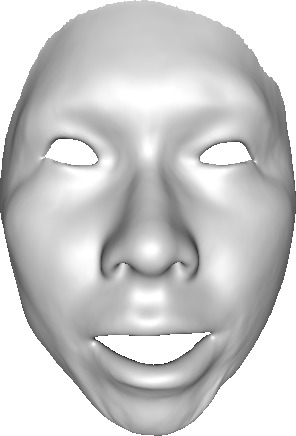}&
	\includegraphics[height=.075\textheight]{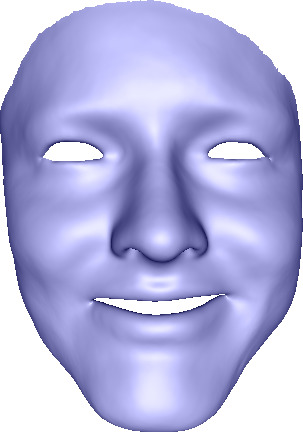}
	\includegraphics[height=.075\textheight]{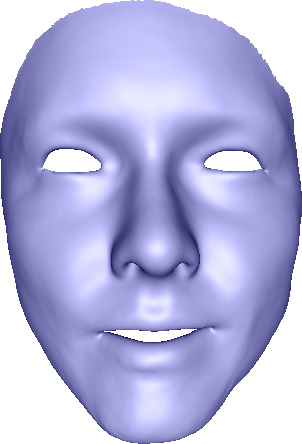}
	\includegraphics[height=.075\textheight]{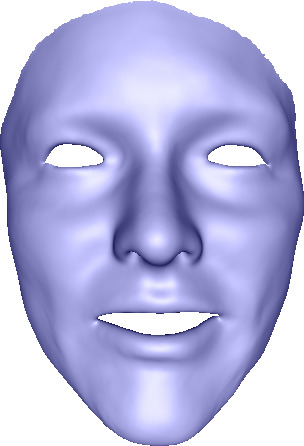}
	\includegraphics[height=.075\textheight]{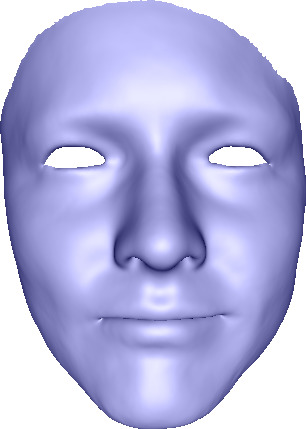} \\
&
\begin{subfigure}[b]{0.3\textwidth}
\caption{Without feature loss}
\label{fig:ablation-no-feat}
\end{subfigure}\vspace{-5pt}
&\\
	\includegraphics[height=.075\textheight]{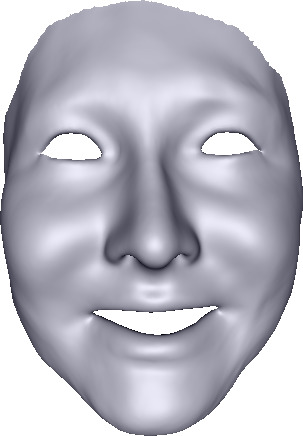}
	\includegraphics[height=.075\textheight]{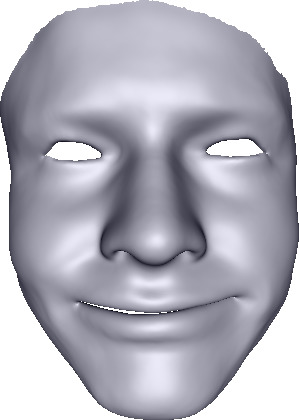}
	\includegraphics[height=.075\textheight]{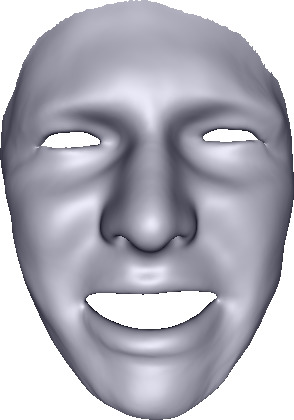}
	\includegraphics[height=.075\textheight]{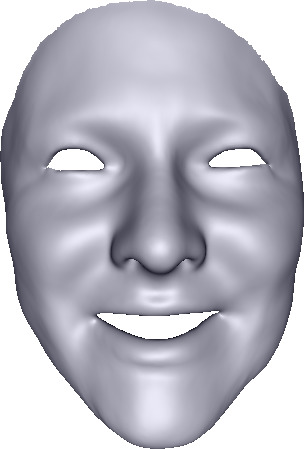} &
	\includegraphics[height=.075\textheight]{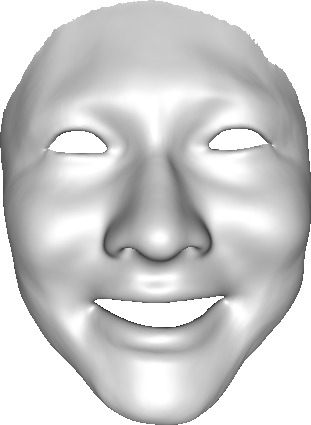}
	\includegraphics[height=.075\textheight]{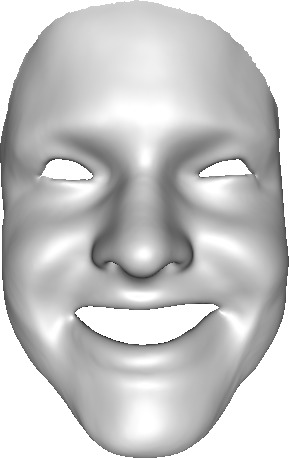}
	\includegraphics[height=.075\textheight]{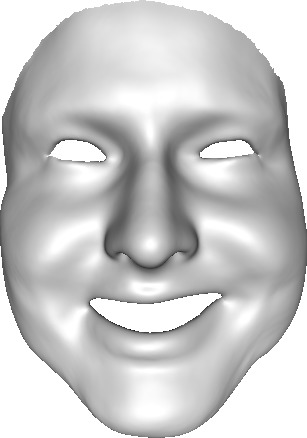}
	\includegraphics[height=.075\textheight]{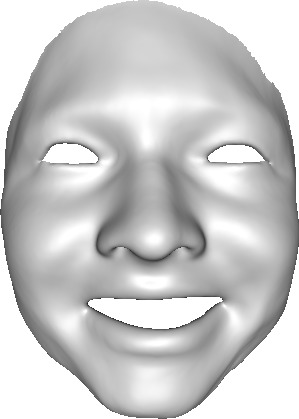}&
	\includegraphics[height=.075\textheight]{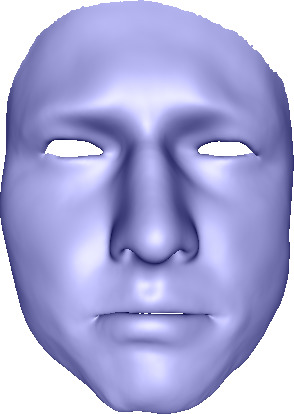}
	\includegraphics[height=.075\textheight]{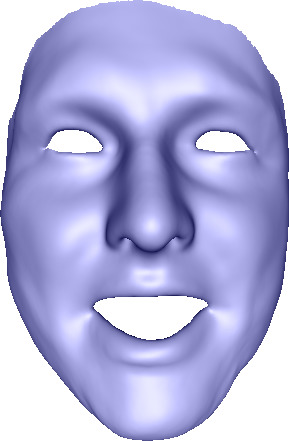}
	\includegraphics[height=.075\textheight]{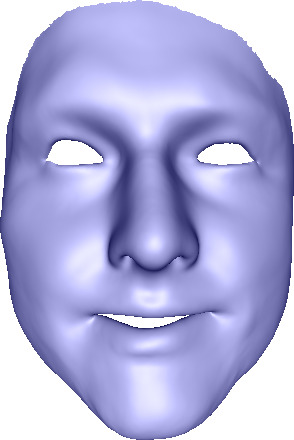}
	\includegraphics[height=.075\textheight]{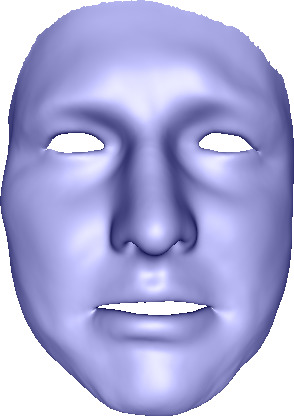}\\
&
\begin{subfigure}[b]{0.3\textwidth}
\caption{Proposed}
\label{fig:ablation-proposed}
\end{subfigure}\vspace{-10pt}
&
\end{tabular}
\caption{Qualitative results for alternative approaches. From left to right: randomly generated samples (dark gray), random samples with a same expression code (light gray), random samples with a same identity code (purple).}
\label{fig:ablation}
\end{figure*}

We start by demonstrating that each of the proposed components is necessary to obtain state-of-the-art results in the proposed metrics.
To this end, we compare our approach against the following alternatives: (1) without mesh regularization (Equation \ref{eq:reg}); (2) with identity classification only; (3) with expression classification only; and (4) without feature loss (Equations \ref{eq:feat_loss_id} and \ref{eq:feat_loss_exp}).

Table \ref{tab:quantitative} gives the evaluation metrics for each of these options, and Figure \ref{fig:ablation} provides qualitative examples. From the results we observe that: (1) The mesh regularization is crucial to generate samples that are realistic facial shapes. This is reflected by a very large value in specificity as well as low diversity, due to the fact that the model never converged to  realistic faces (see Figure~\ref{fig:ablation-no-reg}). (2) Considering classification in only one factor significantly reduces the capacity of the model to preserve semantic properties in the other factor, as indicated by the very low decoupling values obtained in the corresponding rows. This justifies the use of classifiers for each of the factors. (3) Without the feature loss the model can still achieve good results, but both expression decoupling and diversity are lower than with the full model and the inclusion of the feature loss improves expression classification by almost $10\%$. Note that decoupling the  expression space is significantly more challenging than identity, as the provided labels are very sparse. This effect is illustrated on Figure~\ref{fig:ablation-no-feat}, where models with the same expression code can lead to faces with slightly different expressions. Our approach provides more coherent faces,  as shown in Figure~\ref{fig:ablation-proposed}. 


\begin{table}[t]
\begin{center}

\setlength{\tabcolsep}{0.10em} 

\begin{tabular}{|l| c|c|c| c|c|c|}
\cline{2-7}
\multicolumn{1}{l|}{}
 & \multicolumn{1}{c|}{Dec-Id} & \multicolumn{1}{c|}{Dec-Exp} & \multicolumn{1}{c|}{Div} & \multicolumn{1}{c|}{Div-Id} & \multicolumn{1}{c|}{Div-Exp} & \multicolumn{1}{c|}{Sp.}\\
\hline
Training data &  
$-$ & $-$ & $4.89$ & $3.30$ & $5.04$ & $-$\\
\hline

w/o mesh reg. &  
$99.6$ & $99.1$ & $1.41$ & $0.65$ & $1.25$ & $3.61$ \\

w/o exp. class. &  
$100.0$ & $42.8$ & $4.81$ & $0.11$ & $4.87$ & $2.01$ \\

w/o id. class. &  
$7.8$ & $98.9$ & $5.28$ & $4.87$ & $2.05$ & $2.22$ \\

w/o feat. loss &  
$96.0$ & $80.3$ & $4.47$ & $1.75$ & $4.01$ & $2.00$ \\

\hline

3DMM \cite{Amberg08} &
$99.6$ & $65.6$ & $3.53$ & $1.95$ & $2.89$ & $2.30$ \\

MAE \cite{Abrevaya18} &  
$99.5$ & $53.3$ & $3.89$ & $0.92$ & $3.76$ & $2.00$ \\

CoMA \cite{Ranjan18} &  
$97.5$ & $65.5$ & $3.38$ & $1.71$ & $2.90$ & $2.47$ \\

Ours &  
$98.6$ & $89.7$ & $4.74$ & $1.94$ & $4.22$ & $2.01$ \\
\hline 

\end{tabular}

\caption{Quantitative evaluation with respect to decoupling of identity and expression (\emph{Dec-}, percentage), diversity (\emph{Div-}, in mm) and specificity (\emph{Sp.}, in mm.). Higher is better, except for specificity.}
\label{tab:quantitative}

\end{center}
\end{table}

\subsection{Comparisons}
\label{sec:comparison}

\begin{figure}[t]
\centering

\begin{subfigure}[b]{0.45\textwidth}\vspace{10pt}
\begin{tabular}{c c c c c}
Source & Target & CoMA & MAE & Ours\\
	\includegraphics[height=.069\textheight]{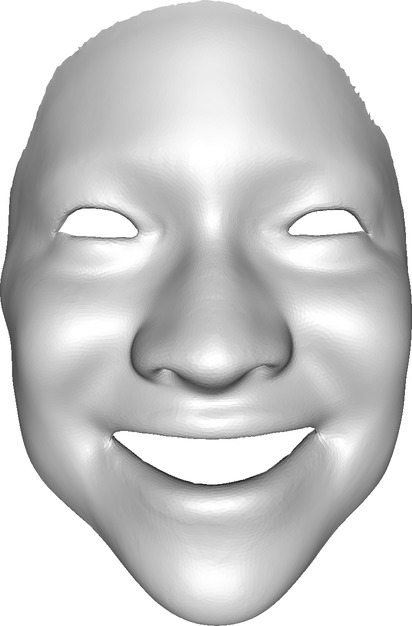} &
	\includegraphics[height=.069\textheight]{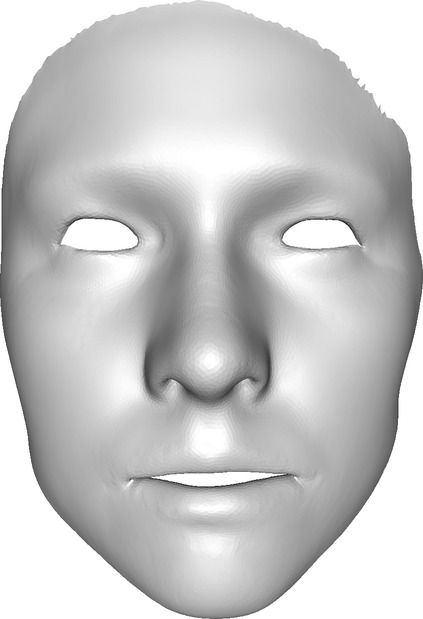} &
	\includegraphics[height=.069\textheight]{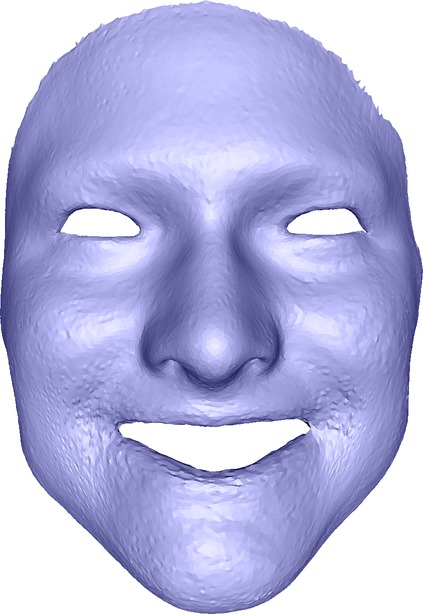} &
	\includegraphics[height=.069\textheight]{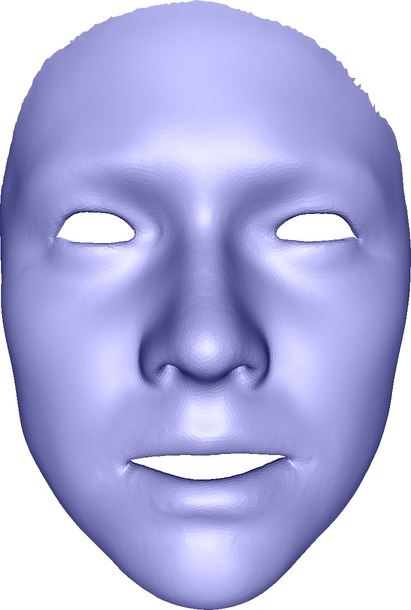} &
	\includegraphics[height=.069\textheight]{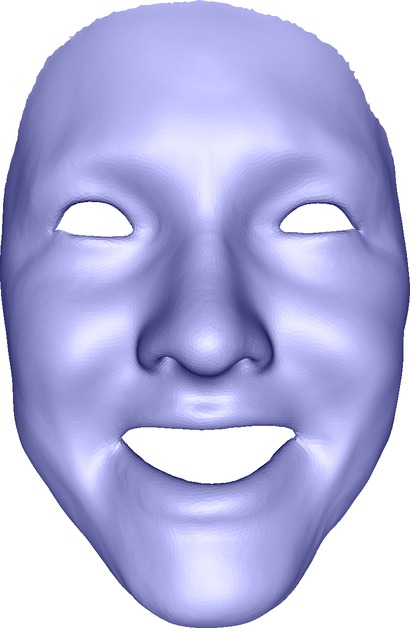}\\
	\includegraphics[height=.069\textheight]{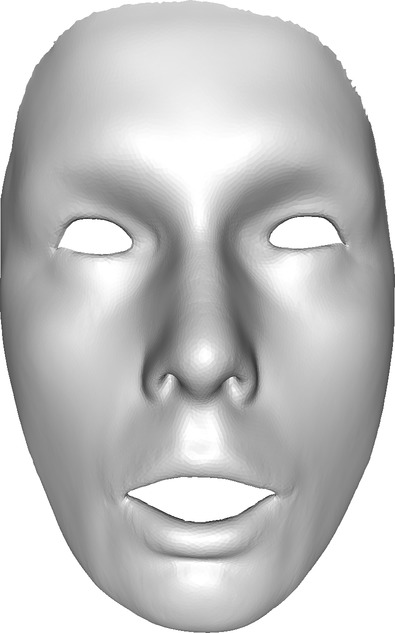} &
	\includegraphics[height=.069\textheight]{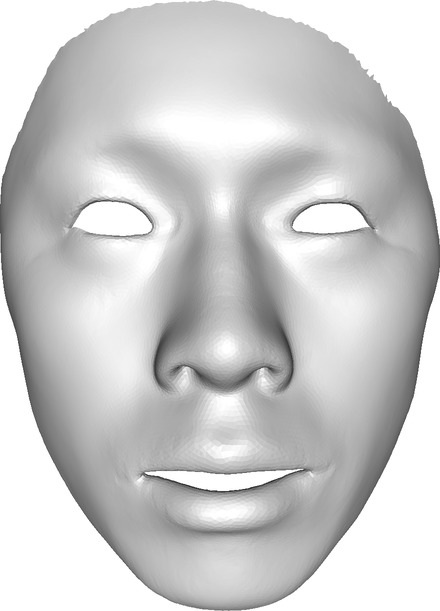} &
	\includegraphics[height=.069\textheight]{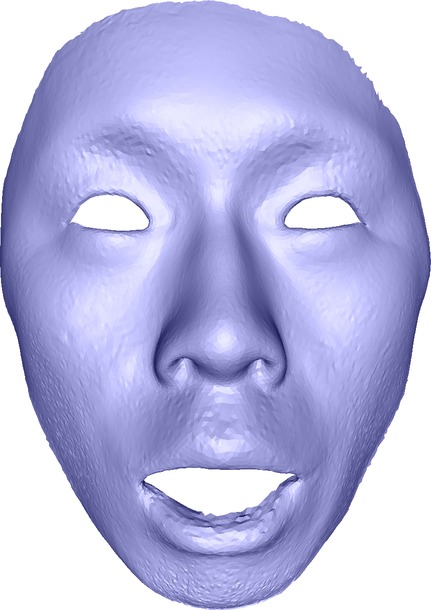} &
	\includegraphics[height=.069\textheight]{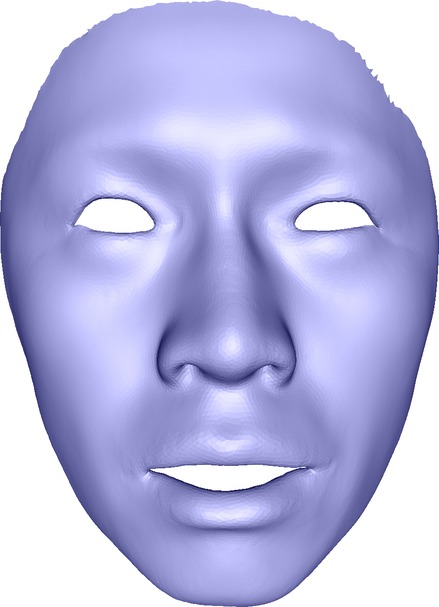} &
	\includegraphics[height=.069\textheight]{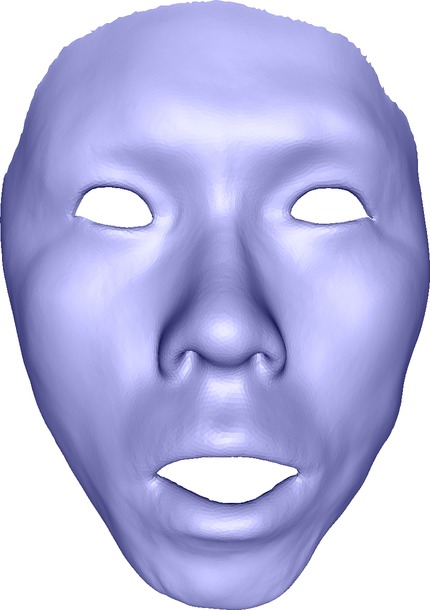}\\
\end{tabular}
\caption{Transferring expression to a target face}
\end{subfigure}%

\begin{subfigure}[b]{0.445\textwidth}
	\includegraphics[height=.069\textheight]{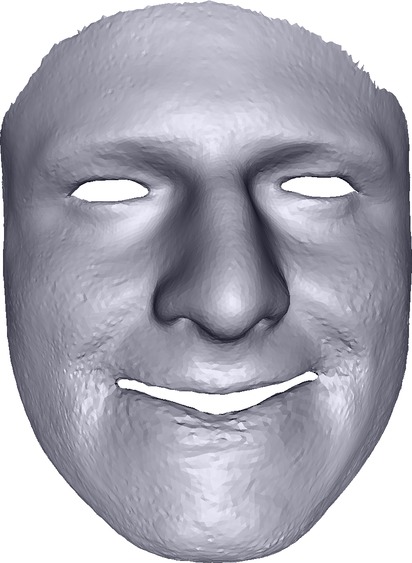}
	\includegraphics[height=.069\textheight]{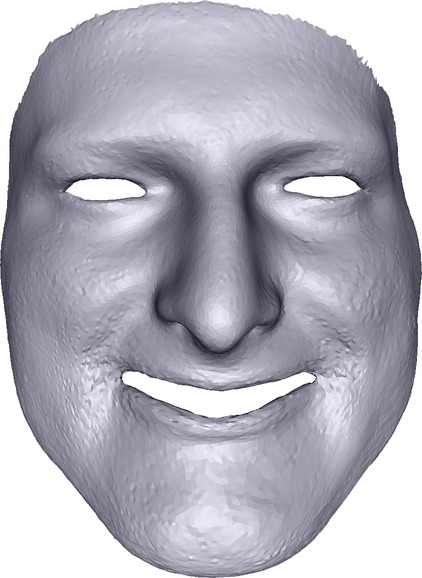}
	\quad
	\includegraphics[height=.069\textheight]{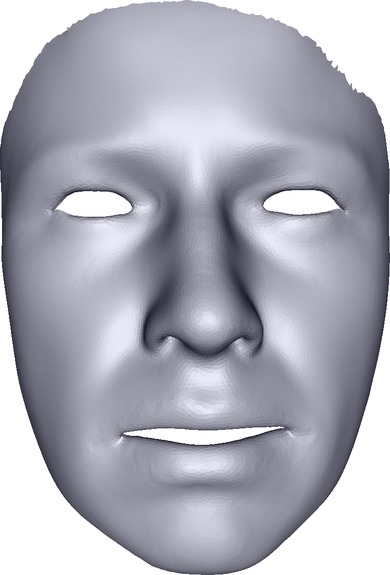}
	\includegraphics[height=.069\textheight]{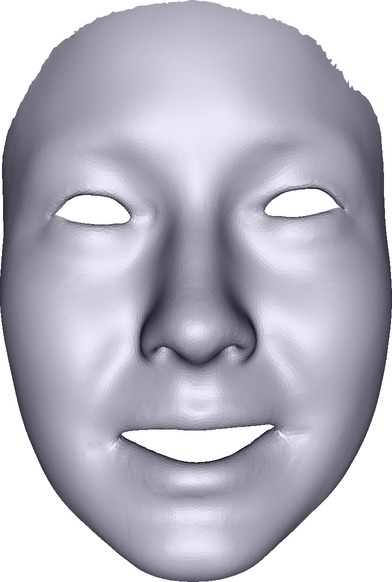}
	\quad
	\includegraphics[height=.069\textheight]{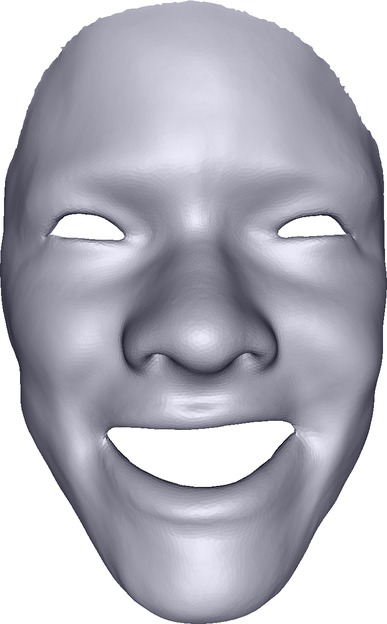}
	\includegraphics[height=.069\textheight]{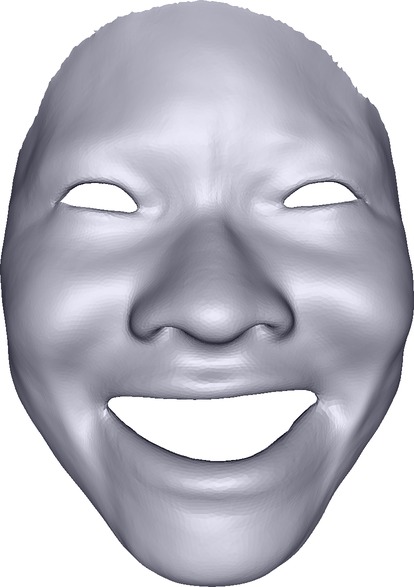}	
\caption{Sampling novel identities from the transferred expression.}
\end{subfigure}
\caption{Comparison in terms of expression transfer. Top: expression code $z_{expr}$ transferred to a target identity. Bottom: using $z_{expr}$ from the source in the top row, we sample novel identities (left to right: CoMA, MAE, ours).}
\label{fig:comparison}
\end{figure}

We compare the proposed approach against state-of-the-art generative 3D face models. Our goal is to build a decoupled latent space, and thus we focus the comparison to works that either enforce this explicitly \cite{Abrevaya18}, or combine a model trained on expressions with a linear space of identities \cite{Ranjan18, Amberg08}. We train all models using the same dimensions ($65$ for identity and $20$ for expression).

The model proposed in \cite{Abrevaya18}, called MAE in the following, was trained with the same dataset and the same label information (Section~\ref{sec:datasets}) for $200$ epochs, with the default parameters given in the paper. We initialize the encoder and the decoder from the publicly available models.

The model proposed in \cite{Ranjan18}, called CoMA in the following, does not explicitly favor decoupling and thus we use the DeepFLAME alternative~\cite{Li2017}, which we also train with the same dataset. This results in a PCA model built from $299$ identities and an autoencoder trained on $30330$ displacements from the corresponding neutral face. For the identity space we manually selected one neutral frame for each sequence in BP4D-Spontaneous, as this dataset does not provide labels. The model was trained using the publicly available code for $200$ epochs.

We also trained an additive linear model as described in~\cite{Amberg08} using our dataset, and the same neutral/expression separation selected for CoMA (see above). We refer to this model as 3DMM.

\myparagraph{Model quality} We show quantitative results with respect to decoupling, diversity and specificity in the bottom of Table~\ref{tab:quantitative}. Note that the proposed approach significantly outperforms the others in terms of \emph{expression decoupling}, which is more challenging than identity due to the sparse labeling. This is shown qualitatively in Figure~\ref{fig:comparison}, where we transferred expressions by simply exchanging the latent code $z_{exp}$. We can see here that the expression is well preserved by our model. 

With respect to \emph{identity decoupling} the four methods perform similarly well, with 3DMM achieving the highest value. Note that, in the case of MAE, the large decoupling value is combined with the lowest diversity in identity (\emph{Div-Id}), which suggests limited generative capabilities (see supplemental for a qualitative example). 

We also outperform all methods in terms of \emph{diversity}. Combined with a specificity value that is among the best, this implies that our model has learned to generate significant variations that remain valid facial shapes.

\myparagraph{Reconstruction of Sparse Data}
We also tested the generalization of the model  with the reconstruction of partial face data given very sparse constraints. 
To this purpose, we use the dataset provided by~\cite{Ranjan18}, which contains $12$ subjects performing $12$ extreme expressions. We take the middle frame of each sequence and manually label $85$ landmarks (see supplemental), resulting in a testing set of $144$ subjects.
The face model is fitted by minimizing:
\begin{equation}
\argmin_{z} \sum_{i = 1}^{p} || \mathbf{\tilde{v}}_i(z) - \mathbf{v}_i ||_2^2 + \lambda ||z||_2^2,
\end{equation}
where $\mathbf{v}_i$ are the 3D locations of the $p$ key-points in the testing set, $\mathbf{\tilde{v}}_i(z)$ are the corresponding key-points in the face model generated with code $z$, and $\lambda$ the regularization weight. We optimize using a gradient descent approach \cite{Kingma14} starting from a randomly sampled code $z$.
Note that this is a  challenging scenario since the training set does not contain such expressions, and the correspondences are very sparse.

We compare our results with those obtained with 3DMM, MAE and CoMA, using the same optimization for all methods. We measure the reconstruction error against the ground-truth surface and report the RMSE. Quantitative results can be found in Table~\ref{tab:reconstruction} for different regularization weights $\lambda$. Our method outperforms in all cases, including  without regularization ($\lambda=0$). We found that our model can produce reasonable faces in most cases, while MAE and CoMA easily produce un-realistic faces when the regularization is not strong enough  (qualitative examples can be found in the supplemental material). 

\begin{table}[t]
\begin{center}
\begin{tabular}{|l| c|c|c|}
\hline
Method & $\lambda=0$ & $\lambda=0.01$ & $\lambda=10$\\
\hline
3DMM~\cite{Amberg08} & $6.62$ & $4.64$ & $2.46$\\
MAE~\cite{Abrevaya18} & $4.46$ & $4.06$ & $2.78$ \\
CoMA~\cite{Ranjan18} & $3.05$ & $3.02$ & $2.83$\\
Ours & $\mathbf{2.62}$ & $\mathbf{2.55}$ & $\mathbf{2.42}$\\
\hline 
\end{tabular}
\end{center}
\caption{Reconstruction of sparse data under different regularization weights (RMSE, in mm).}
\label{tab:reconstruction}
\end{table}

\subsection{Extension to other factors} The proposed framework can easily be extended to other factors of variation, such as identity/expression/viseme. We refer to the supplemental material for an example of such a model.

\section{Conclusion}
We explored in this work the use of adversarial training for learning decoupled 3D facial models and showed that we can achieve state-of-the-art performance in terms of decoupling and diversity of the generated samples. This was obtained through a novel 3D-2D architecture, as well as a training scheme that explicitly encourages decoupling through the use of classifiers. Although the expressiveness of the model remains limited by the diversity of the training data and the accuracy of its labels, we show that adversarial learning has the potential to make better use of the available data in building performant 3D facial models. 

{\small
\bibliographystyle{ieee_fullname}
\bibliography{facegan}
}

\begingroup
\let\clearpage\relax 
\onecolumn 
\newpage
\begin{centering}
{\Large \bf A Decoupled 3D Facial Model by Adversarial Training \\ \vspace{5pt} Supplementary Material}\\
\vspace{12pt}
\end{centering}

\section*{Identity-Expression-Viseme Model}
One of the benefits of our framework lies in its ability to easily  extend to other factors of variation.  As an illustration, we trained a model that decouples identity, expression and viseme (the visual counterpart of a phoneme). The results can be found in Figure~\ref{fig:viseme}, where we show qualitative examples obtained by modifying the different factors of variation individually.

We trained the model using the audiovisual 3D dataset of Fanelli \etal~\cite{Fanelli10}, which contains sequences of $14$ subjects performing $40$ speech sequences in neutral and ``expressive'' mode. We assign phoneme labels using the Montreal Forced Aligner tool \cite{McAuliffe17} with the provided audio, which are mapped to visemes following~\cite{Neti00}. For expression, we manually labeled $699$ frames with the aid of the provided expression ratings of each sequence. This resulted in a database with $100\%$ labeled identites, $68\%$ labeled visemes, and $3\%$ labeled expressions. We set the latent dimensions to $(50,50,50,5)$ for identity, expression, viseme and noise, respectively.

Note this is a simplified model of speech, since the temporal information is not taken into account. Yet, we can see in Figure~\ref{fig:viseme} that a decoupling between the aspects affected by phoneme production, and those affected by expressions such as happiness or surprise can be easily distinguished by our framework.
It is also worth noting that these results were obtained with fully automatic labels for viseme, and very sparse manual labels for expression, thus simplifying the efforts required to annotate the dataset.
Unlike the identity and expression factors, which are intuitively easier to separate, the viseme and expression factors are more intertwined and decoupling them is very challenging even for a human annotator. In spite of this, our results show that we can reasonably decouple the three factors.

\begin{figure}[H]
\centering 
\begin{tabular}{c c |  	c c c c c c}
& & \multicolumn{5}{c}{Expression} \\
& & \textbf{Neutral} & \textbf{Disgust} & \textbf{Happy} & \textbf{Sad} & \textbf{Surprise}\\ 
\hline
\multirow{13}{*}{\rotatebox[origin=c]{90}{Phoneme group}}
& \multirow{-2}{*}{\rotatebox[origin=c]{90}{\textbf{/p/ /b/ /m/}}} & 
\includegraphics[height=.105\textheight]{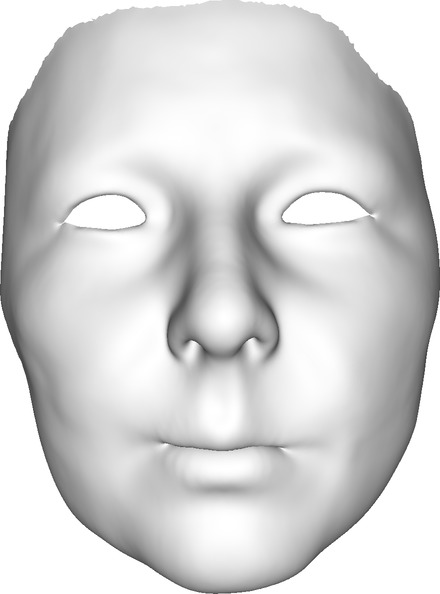} & 
\includegraphics[height=.105\textheight]{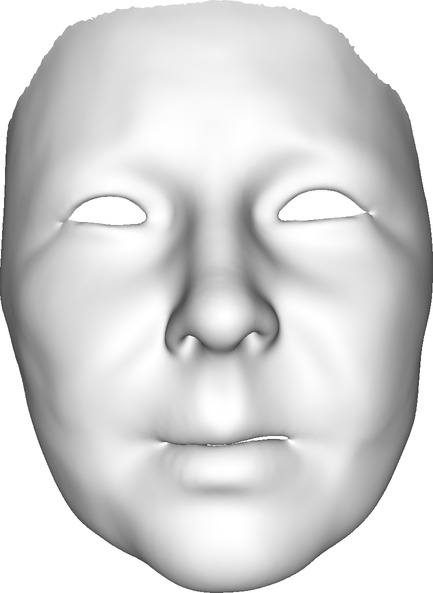} & 
\includegraphics[height=.105\textheight]{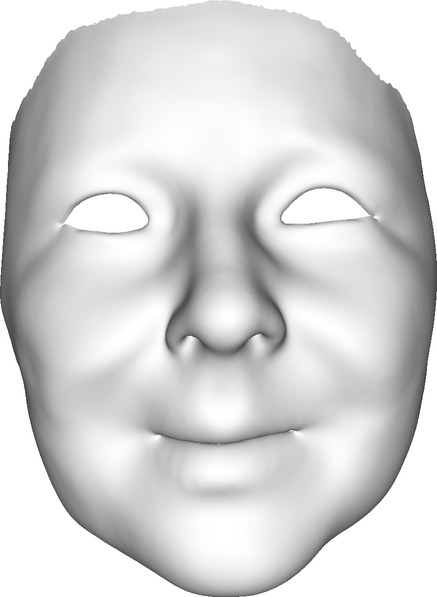} & 
\includegraphics[height=.105\textheight]{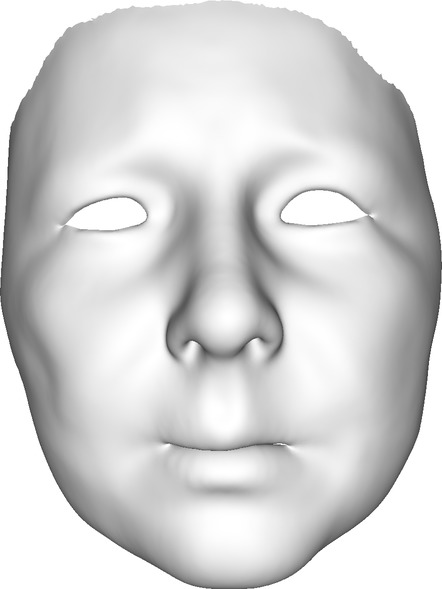} & 
\includegraphics[height=.105\textheight]{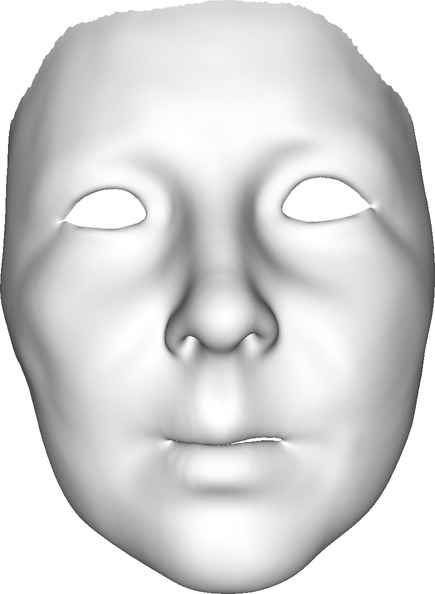}\\
& & 
\includegraphics[height=.105\textheight]{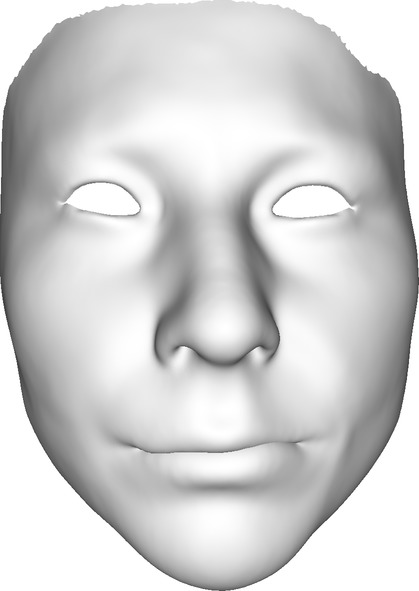} & 
\includegraphics[height=.105\textheight]{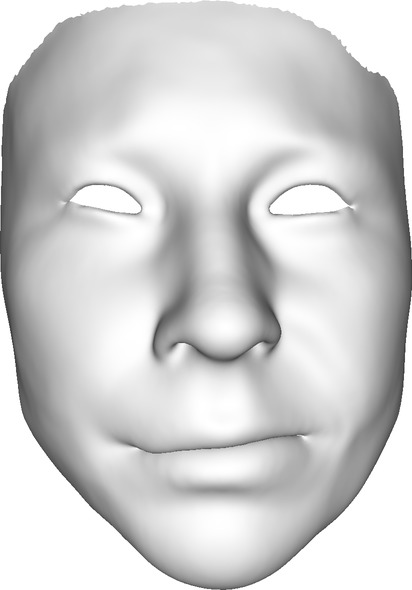} & 
\includegraphics[height=.105\textheight]{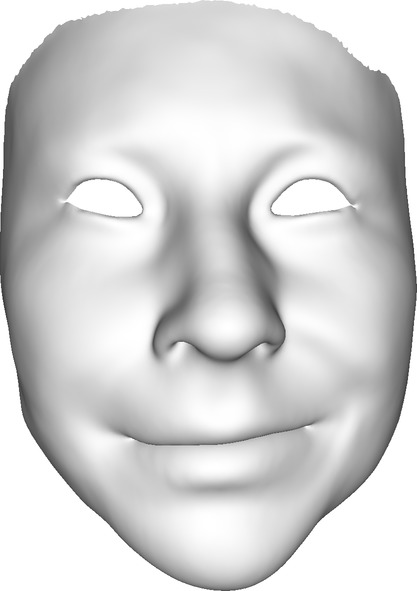} & 
\includegraphics[height=.105\textheight]{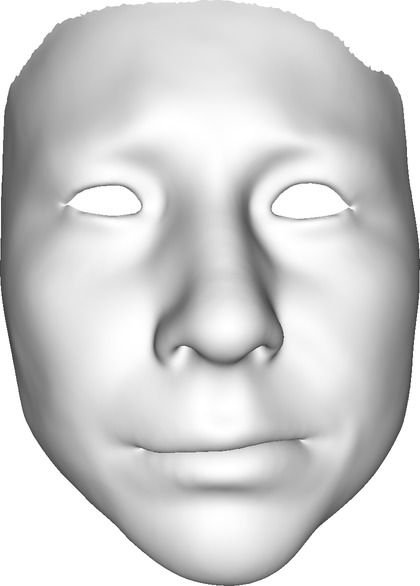} & 
\includegraphics[height=.105\textheight]{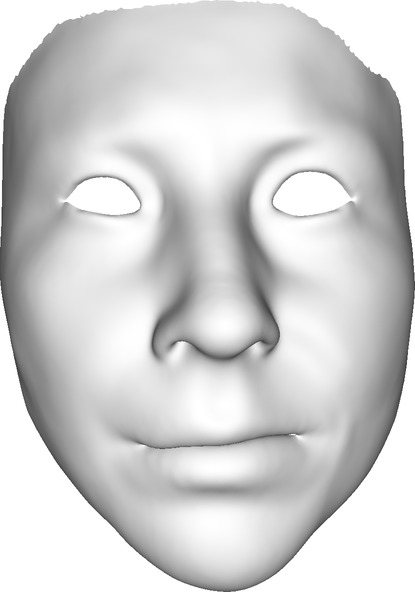}\\ 
\cline{3-7}
& \\
& \multirow{-3}{*}{\rotatebox[origin=c]{90}{\textbf{/uw/ /uh/ /ow/}}} & 
\includegraphics[height=.105\textheight]{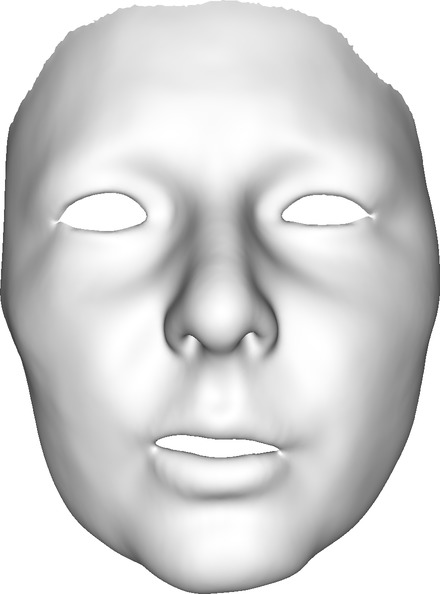} & 
\includegraphics[height=.105\textheight]{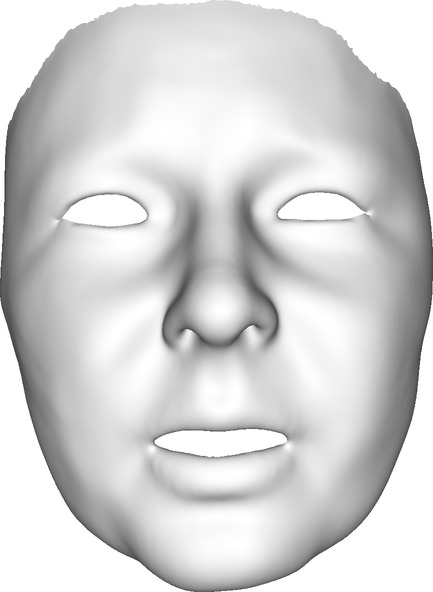} & 
\includegraphics[height=.105\textheight]{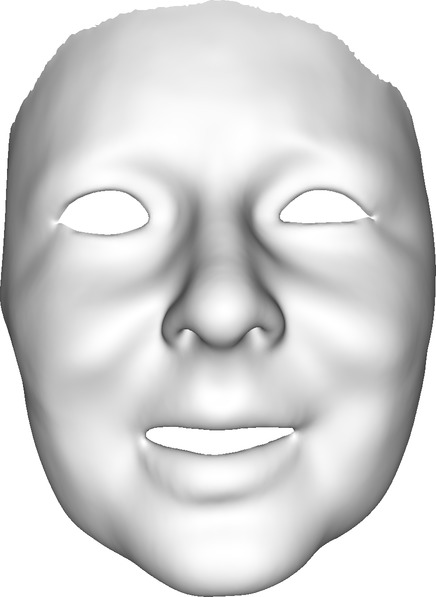} & 
\includegraphics[height=.105\textheight]{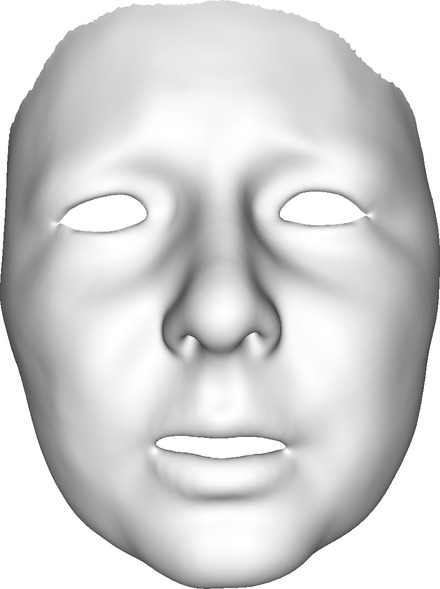} & 
\includegraphics[height=.105\textheight]{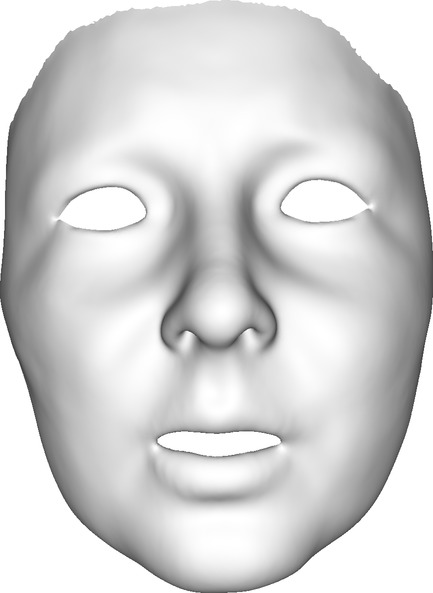}\\
& &
\includegraphics[height=.105\textheight]{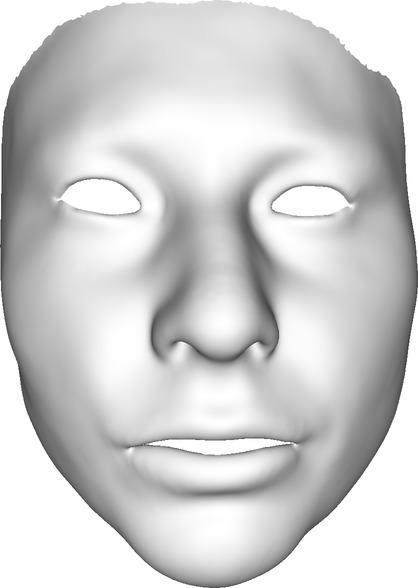} & 
\includegraphics[height=.105\textheight]{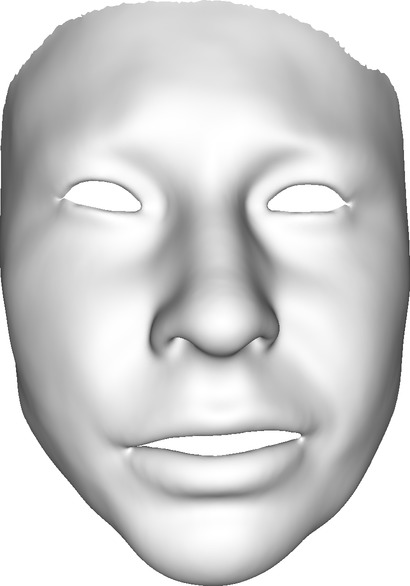} & 
\includegraphics[height=.105\textheight]{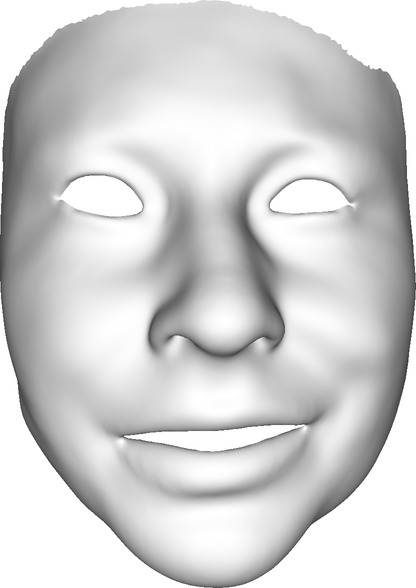} & 
\includegraphics[height=.105\textheight]{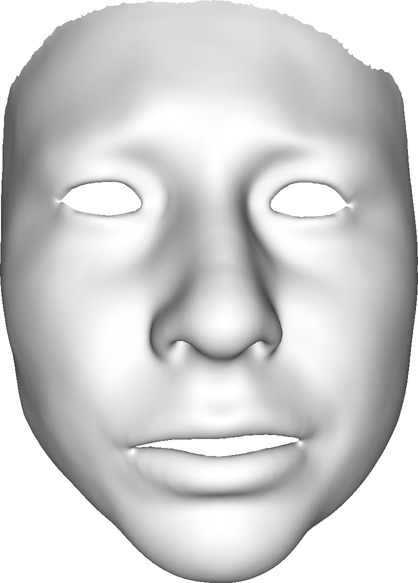} & 
\includegraphics[height=.105\textheight]{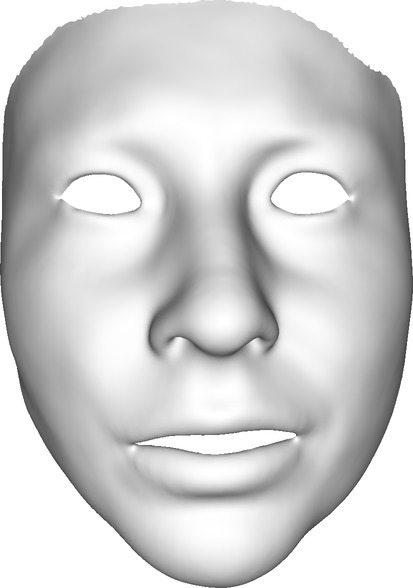}
\end{tabular}
\caption{Example of decoupling between identity, expression and viseme. }
\label{fig:viseme}
\end{figure}

\section*{Latent Space Manipulation}

The following figure shows an example of interpolation and extrapolation in (1) the expression latent space, (2) the identity latent space, and (3) the full latent space:

\begin{figure}[H]
\centering

\begin{tabular}{l l l l l l l l l}
	\includegraphics[height=0.07\textheight]{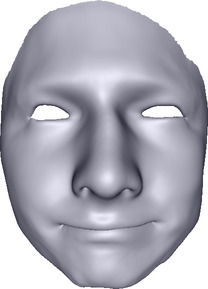} &
	\includegraphics[height=0.07\textheight]{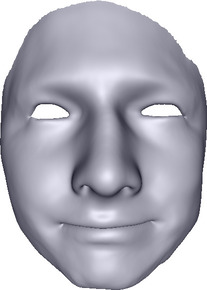} &
	\includegraphics[height=0.07\textheight]{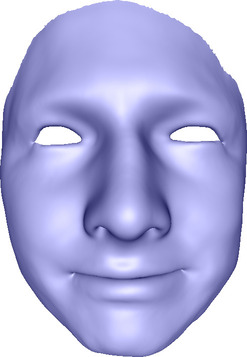} &
	\includegraphics[height=0.07\textheight]{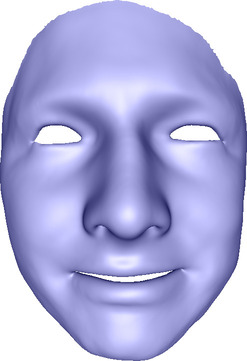} &
	\includegraphics[height=0.07\textheight]{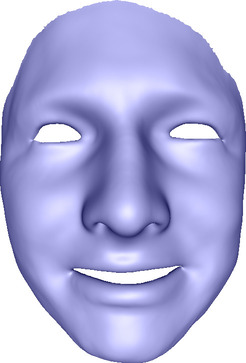} &
	\includegraphics[height=0.07\textheight]{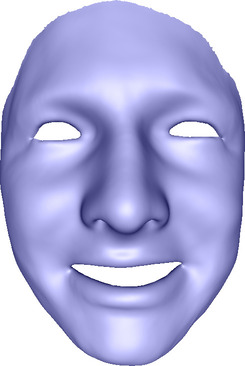} &
	\includegraphics[height=0.07\textheight]{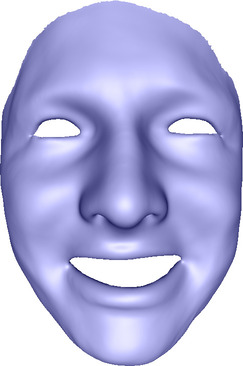} &
	\includegraphics[height=0.07\textheight]{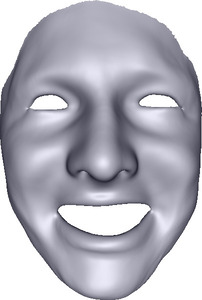} &
	\includegraphics[height=0.07\textheight]{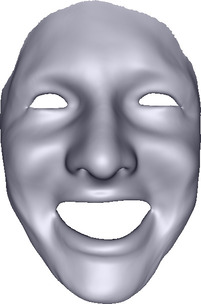}\\
	\includegraphics[height=0.07\textheight]{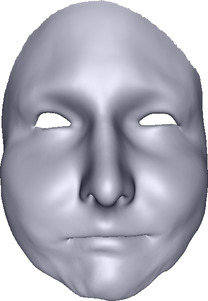} &
	\includegraphics[height=0.07\textheight]{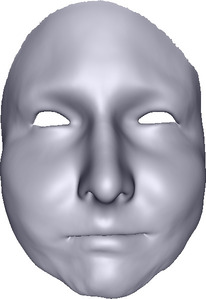} &
	\includegraphics[height=0.07\textheight]{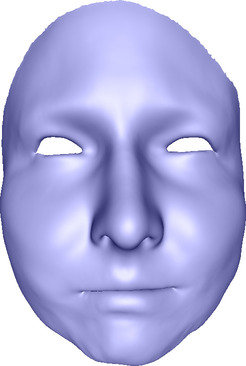} &
	\includegraphics[height=0.07\textheight]{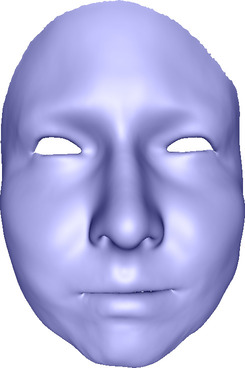} &
	\includegraphics[height=0.07\textheight]{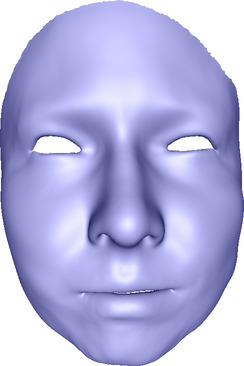} &
	\includegraphics[height=0.07\textheight]{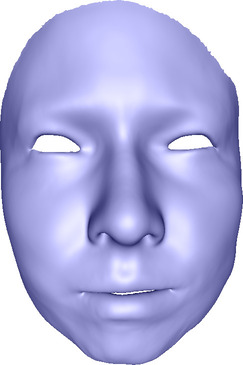} &
	\includegraphics[height=0.07\textheight]{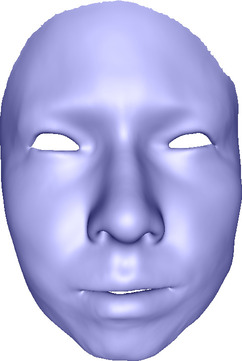} &
	\includegraphics[height=0.07\textheight]{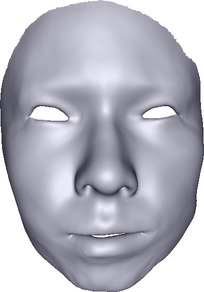} &
	\includegraphics[height=0.07\textheight]{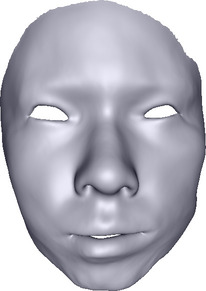}\\
	\includegraphics[height=0.07\textheight]{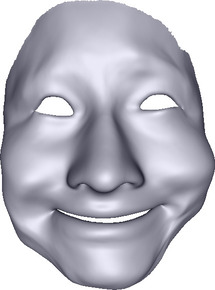} &
	\includegraphics[height=0.07\textheight]{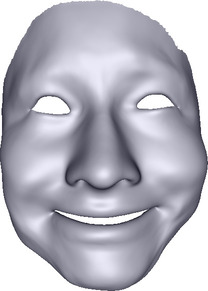} &
	\includegraphics[height=0.07\textheight]{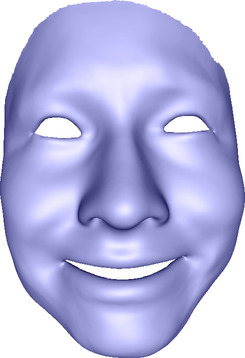} &
	\includegraphics[height=0.07\textheight]{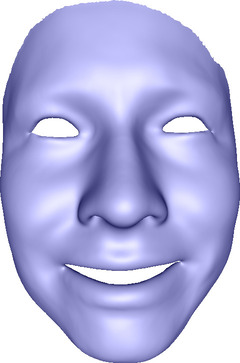} &
	\includegraphics[height=0.07\textheight]{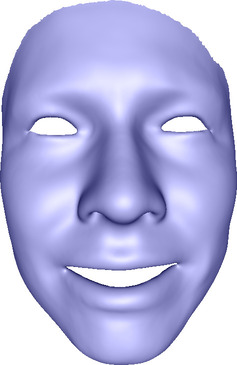} &
	\includegraphics[height=0.07\textheight]{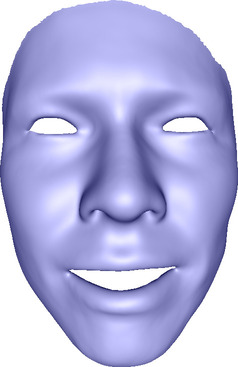} &
	\includegraphics[height=0.07\textheight]{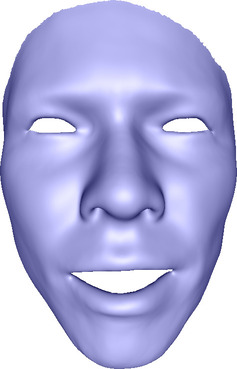} &
	\includegraphics[height=0.07\textheight]{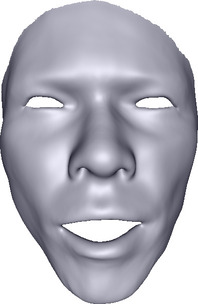} &
	\includegraphics[height=0.07\textheight]{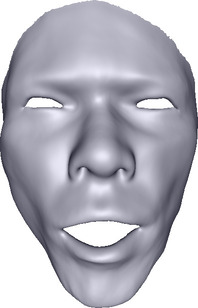}
\end{tabular} 
\caption{From top to bottom: interpolation (purple) and extrapolation (gray) of expression code, identity code, and the full latent.}
\label{fig:int}
\end{figure}

Thanks to the decoupling of identity and expression spaces, we can synthesize new expressions by simple manipulation of the latent space. We show here two possibilities for this.

Given a source mesh obtained with $G(z_{id}^{src}, z_{expr}^{src}, z_{noise}^{src})$ and a target mesh obtained with
$G(z_{id}^{target}, z_{expr}^{target}, z_{noise}^{target})$, we generate new expressions for the target mesh by either

\begin{enumerate}
\itemsep0em
\item Replacing the expression with that of the source: $G(z_{id}^{target}, \mathbf{z_{expr}^{src}}, z_{noise}^{target})$
\item Adding the expression vectors: $G(z_{id}^{target}, \mathbf{z_{expr}^{src} + z_{expr}^{target}}, z_{noise}^{target})$
\end{enumerate}

Results can be seen in Figure~\ref{fig:transfer}. In particular, note how adding the latent vectors results in plausible expressions which preserve the semantics of both sources.
\begin{figure}[H]
\centering
\begin{tabular}{c l c c c l c c c}
\includegraphics[height=.08\textheight]{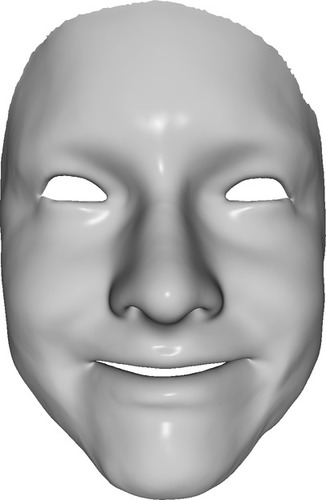} &
\qquad &
\includegraphics[height=.08\textheight]{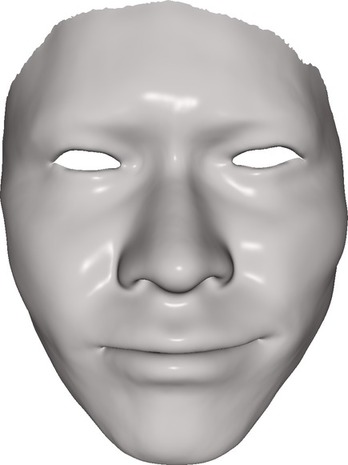} &
\includegraphics[height=.08\textheight]{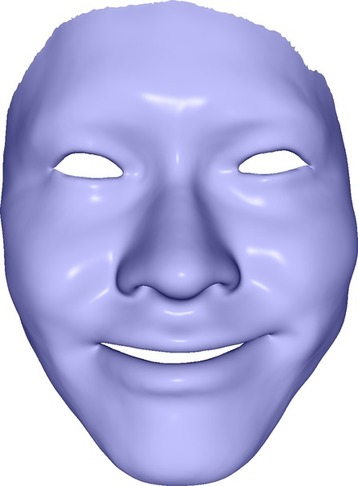} &
\includegraphics[height=.08\textheight]{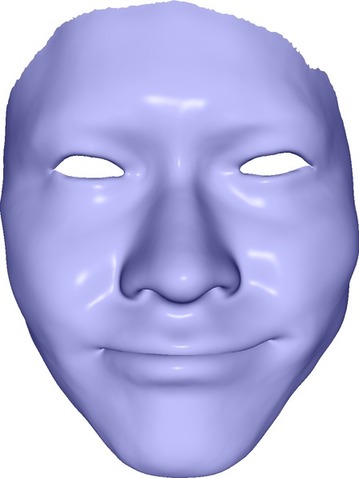} 
& \qquad &
\includegraphics[height=.08\textheight]{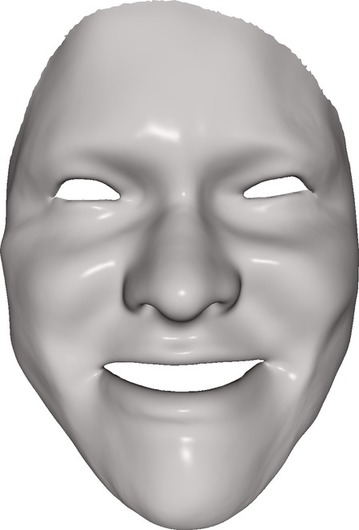} &
\includegraphics[height=.08\textheight]{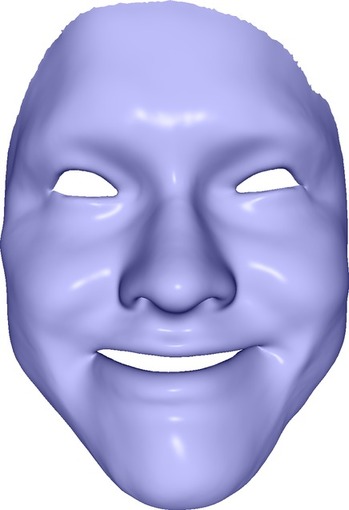} &
\includegraphics[height=.08\textheight]{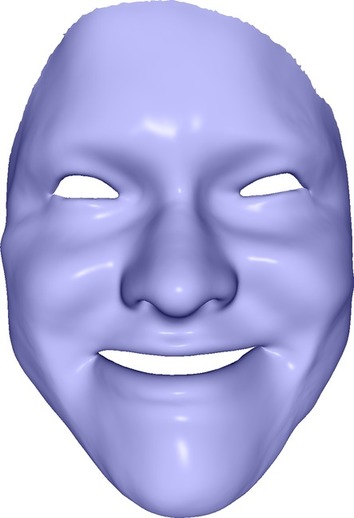}\\
\includegraphics[height=.08\textheight]{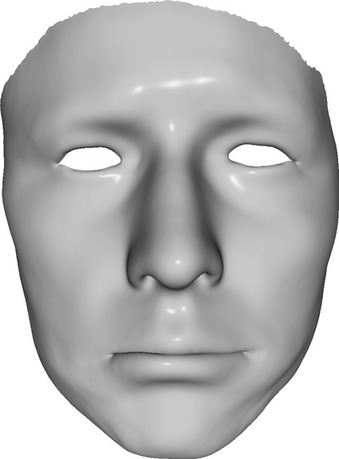} &
\qquad &
\includegraphics[height=.08\textheight]{imgs-supp/transfer-and-sum/targetA} &
\includegraphics[height=.08\textheight]{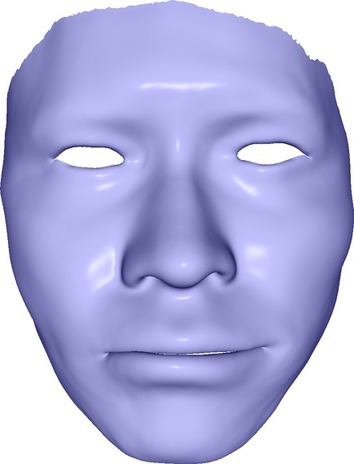} &
\includegraphics[height=.08\textheight]{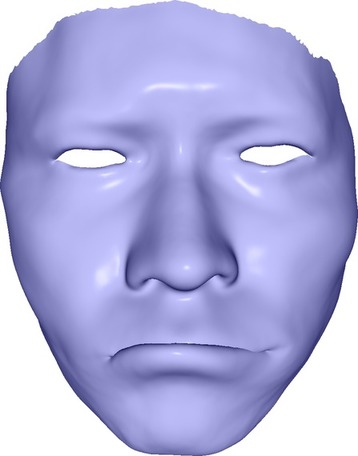} 
& \qquad &
\includegraphics[height=.08\textheight]{imgs-supp/transfer-and-sum/targetC} &
\includegraphics[height=.08\textheight]{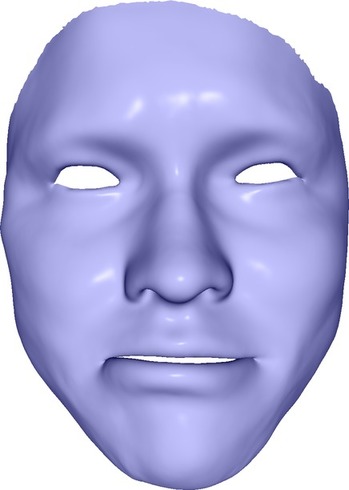} &
\includegraphics[height=.08\textheight]{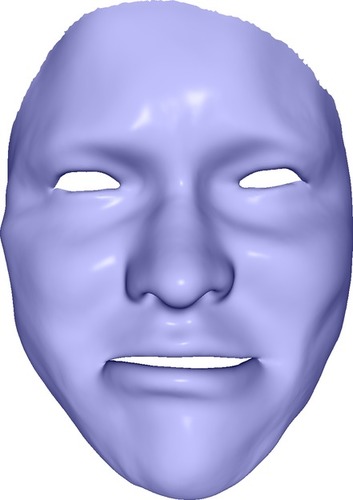}\\
\includegraphics[height=.08\textheight]{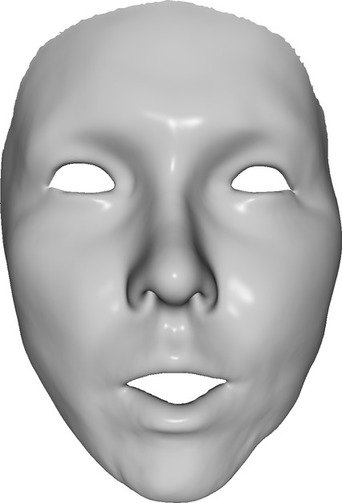} &
\qquad &
\includegraphics[height=.08\textheight]{imgs-supp/transfer-and-sum/targetA} &
\includegraphics[height=.08\textheight]{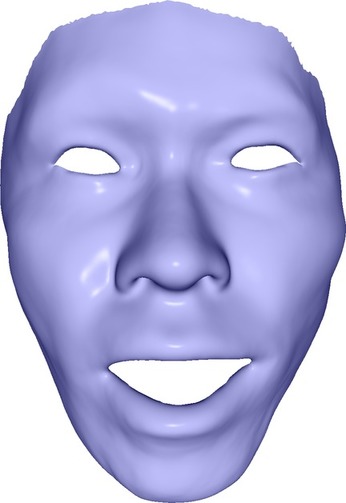} &
\includegraphics[height=.08\textheight]{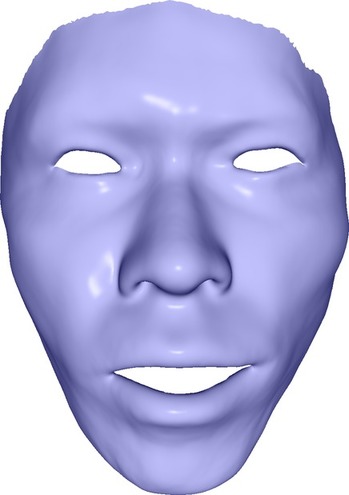} 
& \qquad &
\includegraphics[height=.08\textheight]{imgs-supp/transfer-and-sum/targetC} &
\includegraphics[height=.08\textheight]{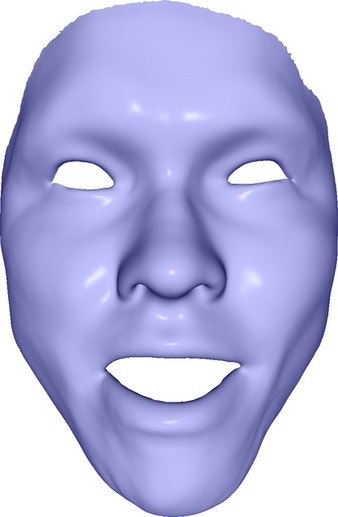} &
\includegraphics[height=.08\textheight]{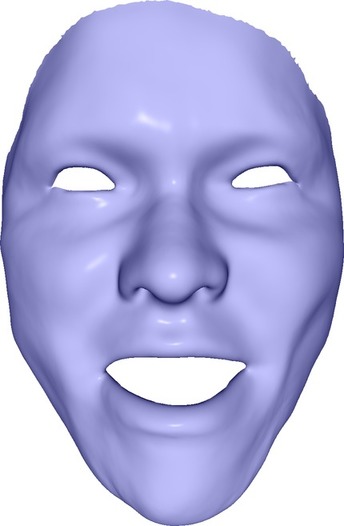}\\
Source & & Target & Replaced & Added & & Target & Replaced & Added
\end{tabular}
\caption{Example of expression space manipulation. In gray a source mesh and a target mesh. In blue the result of (1) replacing the expression code of the target with that of the source \emph{(replaced)}, and (2) adding the source and target expression codes \emph{(added)}.}
\label{fig:transfer}
\end{figure}

\section*{Qualitative Comparisons}

This section provides qualitative examples for the results in Section 5.5, Table 1. Figure~\ref{fig:specificity} shows three random samples with best and worst specificity values, and Figures~\ref{fig:dec_id} and~\ref{fig:dec_expr} show random samples used for decoupling and diversity evaluation of identity and expression, respectively.

\newcolumntype{D}{m{0.085\textwidth}}
\begin{figure}[H]
\centering
\begin{tabular}{c DDD c DDD}
COMA & 
\includegraphics[height=.08\textheight]{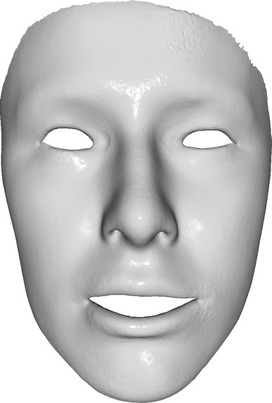} &
\includegraphics[height=.08\textheight]{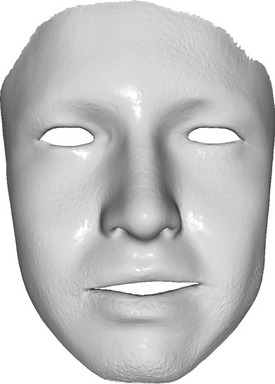} &
\includegraphics[height=.08\textheight]{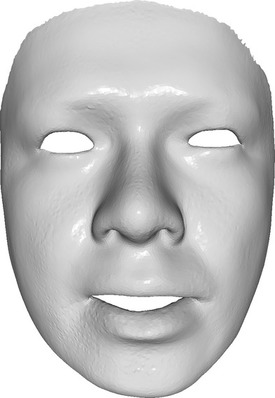} &
\qquad &
\includegraphics[height=.08\textheight]{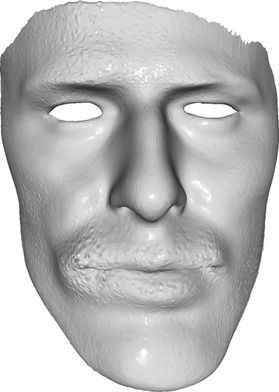} &
\includegraphics[height=.08\textheight]{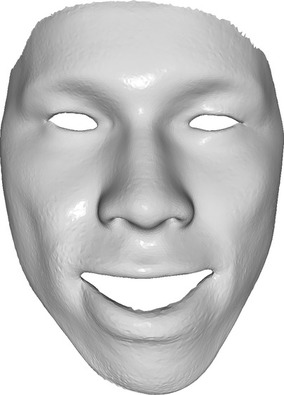} &
\includegraphics[height=.08\textheight]{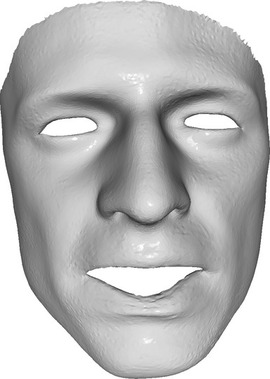} \\
MAE &
\includegraphics[height=.08\textheight]{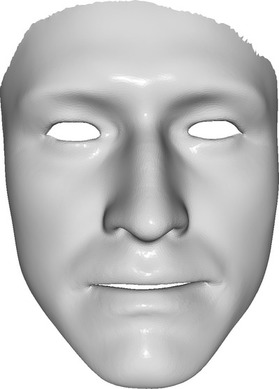} &
\includegraphics[height=.08\textheight]{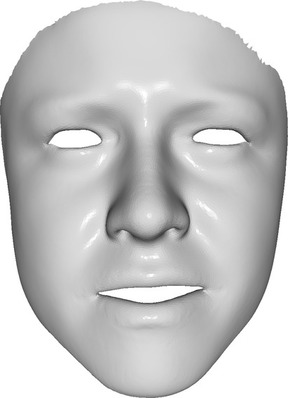} &
\includegraphics[height=.08\textheight]{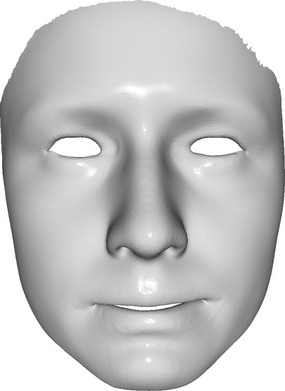} &
\qquad &
\includegraphics[height=.08\textheight]{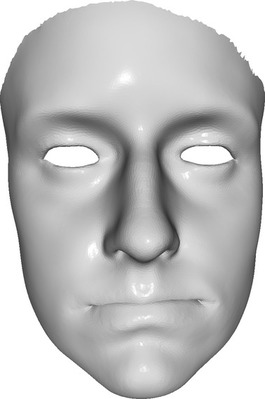} &
\includegraphics[height=.08\textheight]{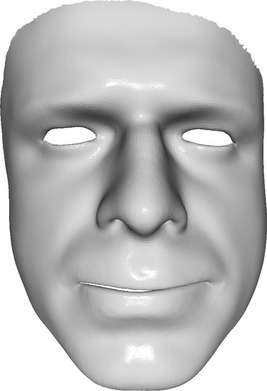} &
\includegraphics[height=.08\textheight]{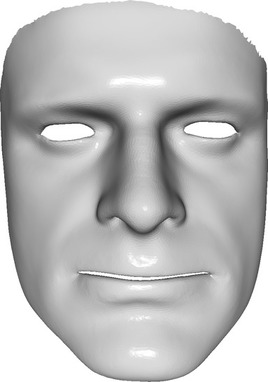} \\
Ours &
\includegraphics[height=.08\textheight]{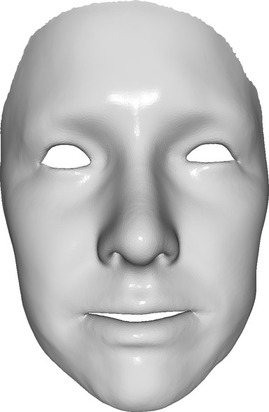} &
\includegraphics[height=.08\textheight]{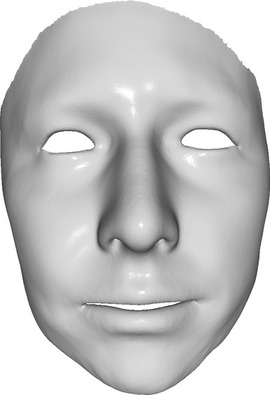} &
\includegraphics[height=.08\textheight]{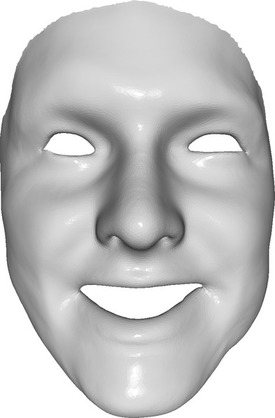} &
\qquad &
\includegraphics[height=.08\textheight]{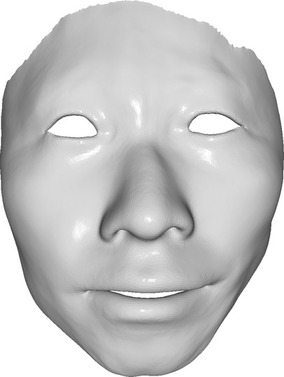} &
\includegraphics[height=.08\textheight]{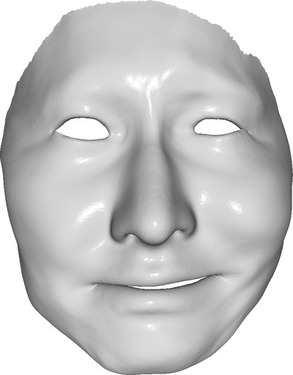} &
\includegraphics[height=.08\textheight]{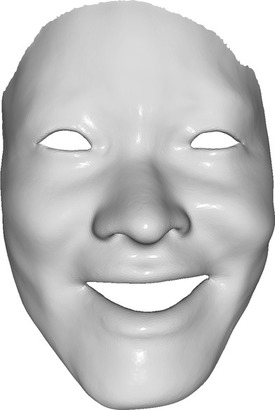}\\
& \multicolumn{3}{c}{Best specificity} & & \multicolumn{3}{c}{Worst specificity} \\
\end{tabular}
\caption{Random samples which obtained the three best (left) and worst (right) values in the specificity metric.}
\label{fig:specificity}
\end{figure}

\newpage
\newcolumntype{A}{m{0.09\textwidth}}
\begin{figure}[H]
\begin{center}

\begin{subfigure}[b]{\textwidth}
\centering
\begin{tabular}{AAAAAAAA}
\includegraphics[height=.085\textheight]{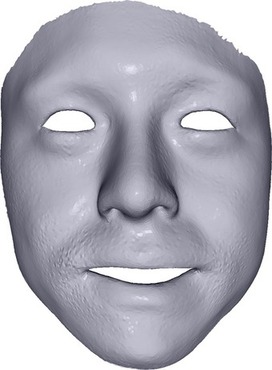} &
\includegraphics[height=.085\textheight]{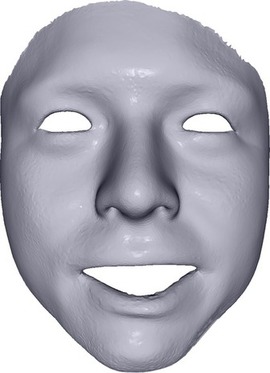} &
\includegraphics[height=.085\textheight]{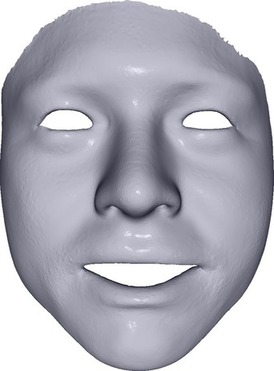} &
\includegraphics[height=.085\textheight]{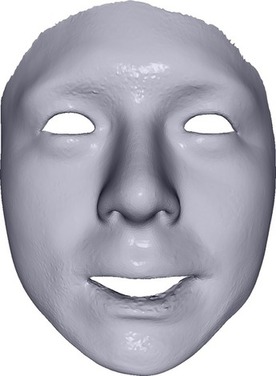} &
\includegraphics[height=.085\textheight]{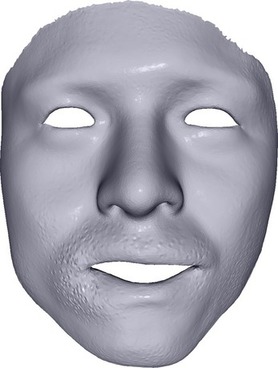} &
\includegraphics[height=.085\textheight]{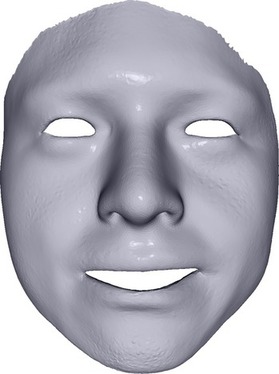} &
\includegraphics[height=.085\textheight]{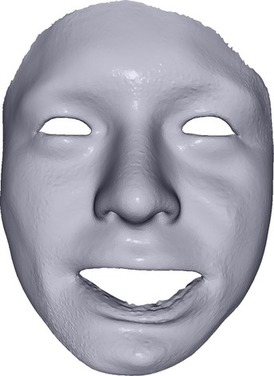} &
\includegraphics[height=.085\textheight]{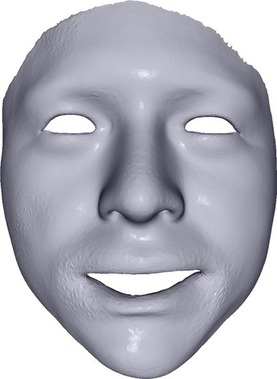}
\\
\includegraphics[height=.085\textheight]{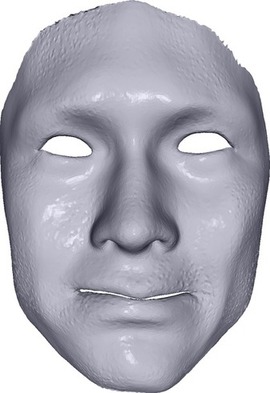} &
\includegraphics[height=.085\textheight]{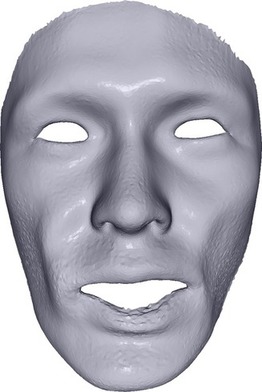} &
\includegraphics[height=.085\textheight]{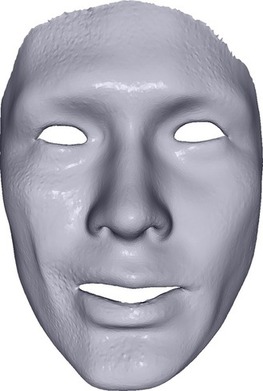} &
\includegraphics[height=.085\textheight]{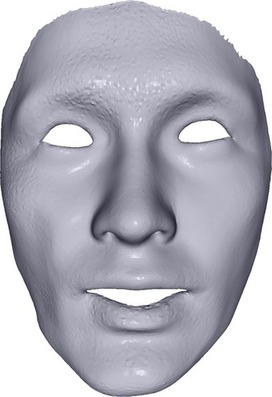} &
\includegraphics[height=.085\textheight]{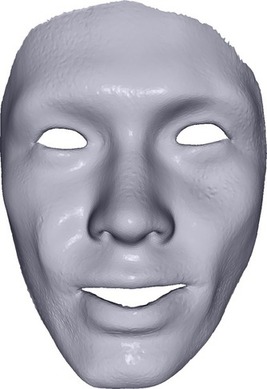} &
\includegraphics[height=.085\textheight]{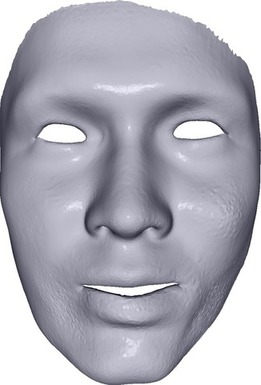} &
\includegraphics[height=.085\textheight]{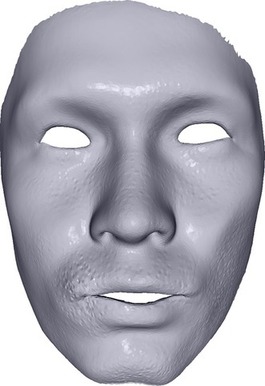} &
\includegraphics[height=.085\textheight]{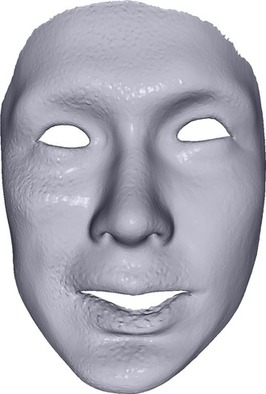}
\\
\includegraphics[height=.085\textheight]{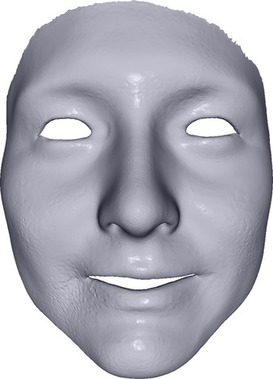} &
\includegraphics[height=.085\textheight]{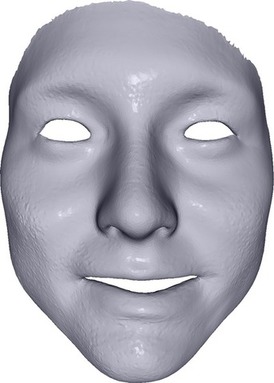} &
\includegraphics[height=.085\textheight]{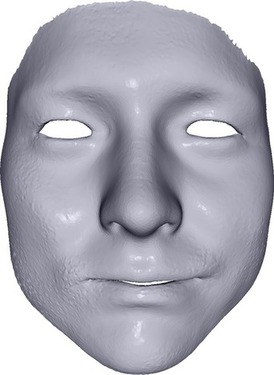} &
\includegraphics[height=.085\textheight]{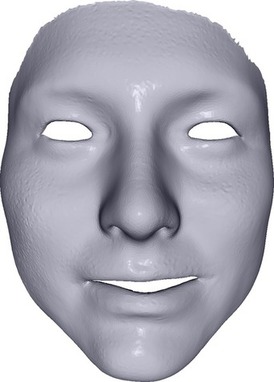} &
\includegraphics[height=.085\textheight]{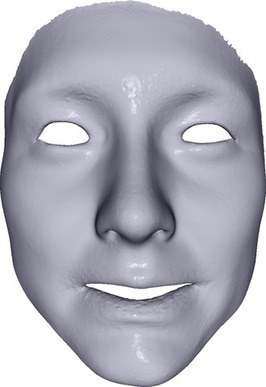} &
\includegraphics[height=.085\textheight]{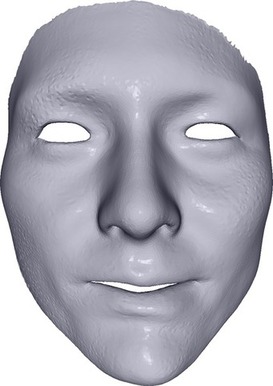} &
\includegraphics[height=.085\textheight]{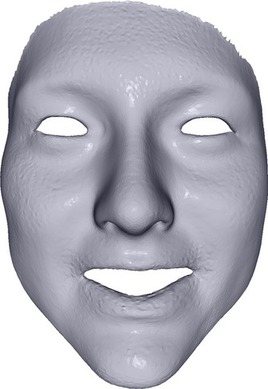} &
\includegraphics[height=.085\textheight]{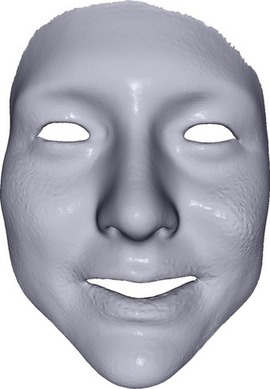}
\end{tabular}
\caption{COMA}
\end{subfigure}

\begin{subfigure}[b]{\textwidth}
\centering
\begin{tabular}{AAAAAAAA}
\includegraphics[height=.085\textheight]{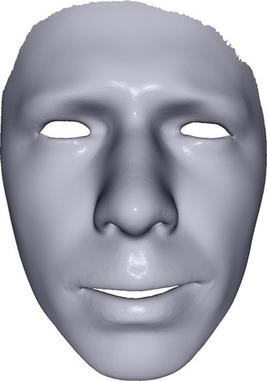} &
\includegraphics[height=.085\textheight]{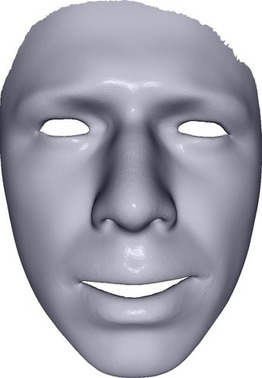} &
\includegraphics[height=.085\textheight]{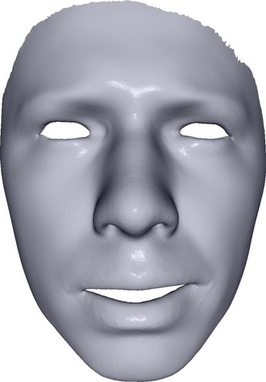} &
\includegraphics[height=.085\textheight]{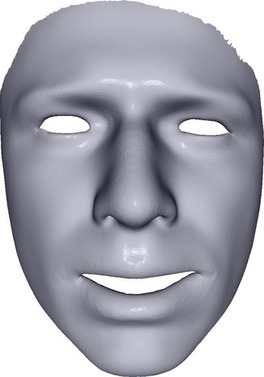} &
\includegraphics[height=.085\textheight]{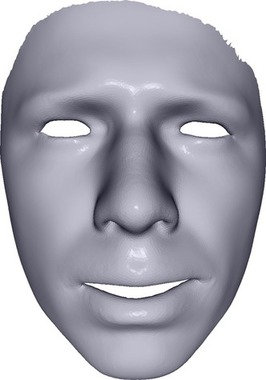} &
\includegraphics[height=.085\textheight]{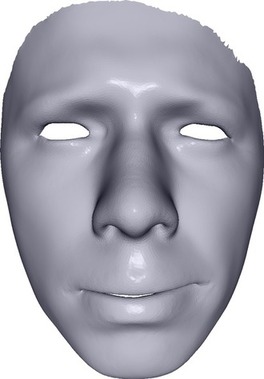} &
\includegraphics[height=.085\textheight]{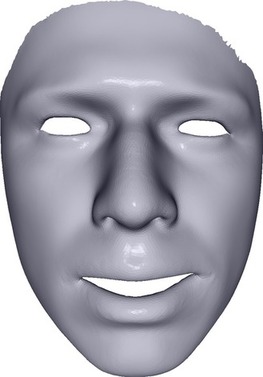} &
\includegraphics[height=.085\textheight]{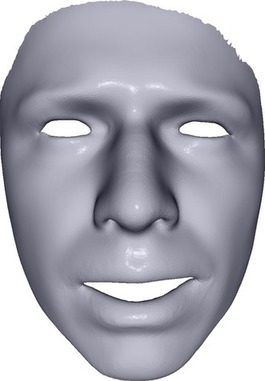}
\\
\includegraphics[height=.085\textheight]{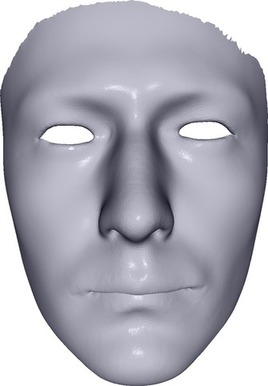} &
\includegraphics[height=.085\textheight]{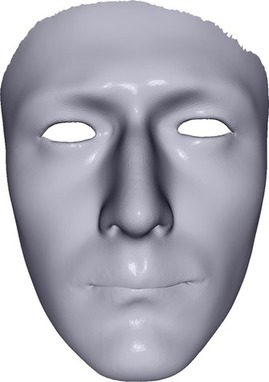} &
\includegraphics[height=.085\textheight]{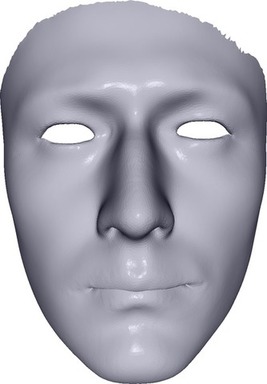} &
\includegraphics[height=.085\textheight]{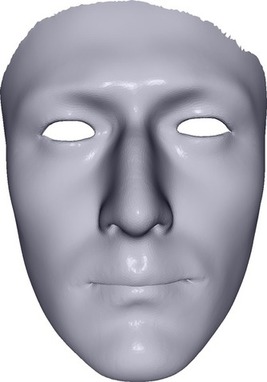} &
\includegraphics[height=.085\textheight]{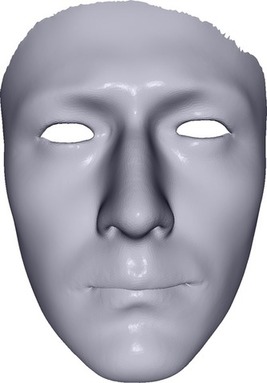} &
\includegraphics[height=.085\textheight]{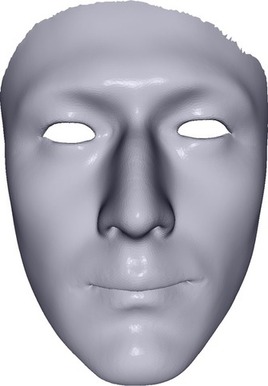} &
\includegraphics[height=.085\textheight]{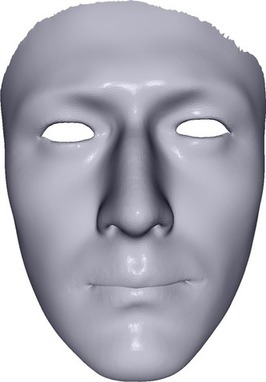} &
\includegraphics[height=.085\textheight]{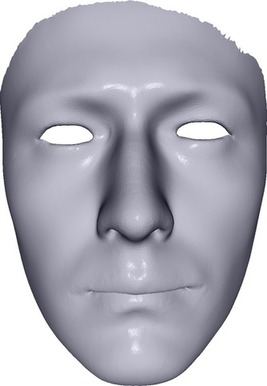}
\\
\includegraphics[height=.085\textheight]{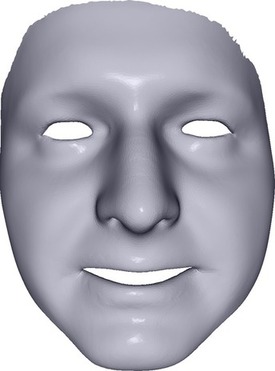} &
\includegraphics[height=.085\textheight]{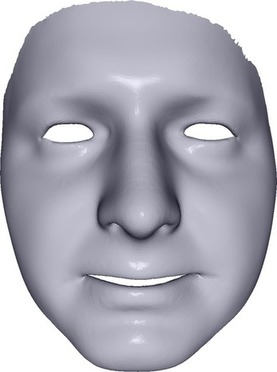} &
\includegraphics[height=.085\textheight]{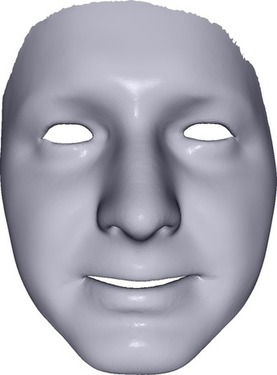} &
\includegraphics[height=.085\textheight]{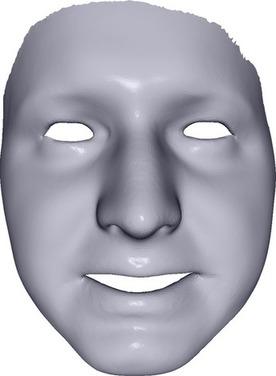} &
\includegraphics[height=.085\textheight]{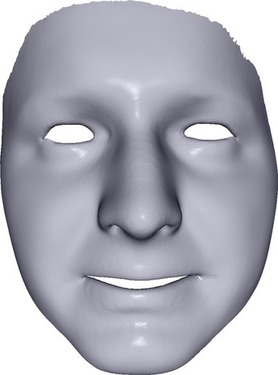} &
\includegraphics[height=.085\textheight]{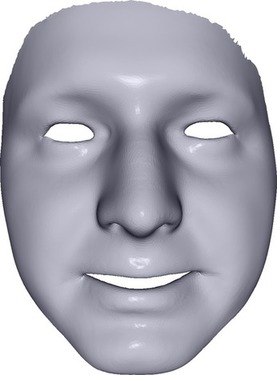} &
\includegraphics[height=.085\textheight]{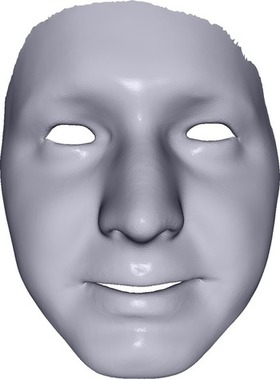} &
\includegraphics[height=.085\textheight]{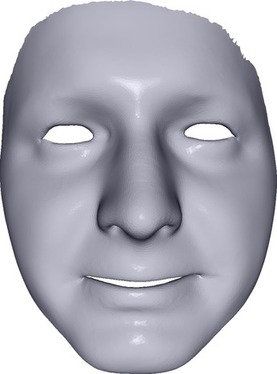}
\end{tabular}
\caption{MAE}
\end{subfigure}

\begin{subfigure}[b]{\textwidth}
\centering
\begin{tabular}{AAAAAAAA}
\includegraphics[height=.085\textheight]{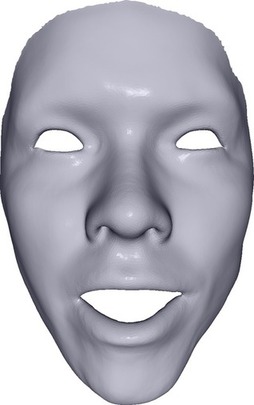} &
\includegraphics[height=.085\textheight]{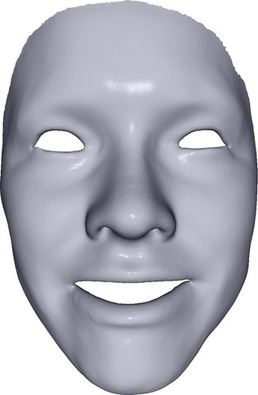} &
\includegraphics[height=.085\textheight]{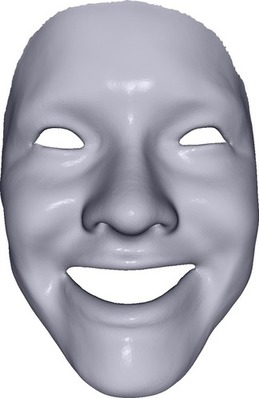} &
\includegraphics[height=.085\textheight]{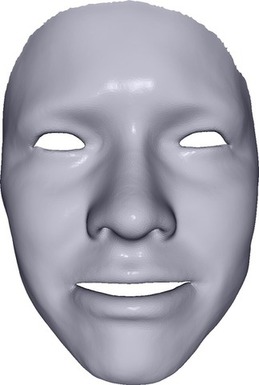} &
\includegraphics[height=.085\textheight]{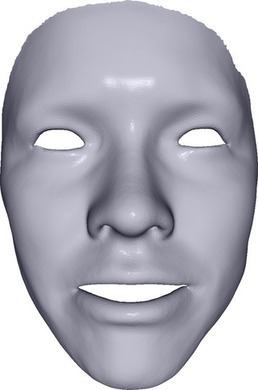} &
\includegraphics[height=.085\textheight]{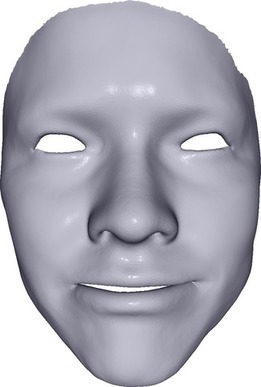} &
\includegraphics[height=.085\textheight]{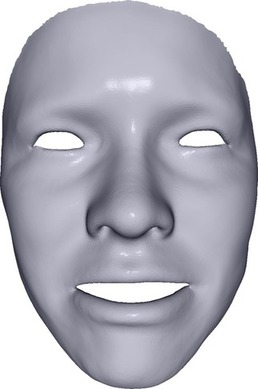} &
\includegraphics[height=.085\textheight]{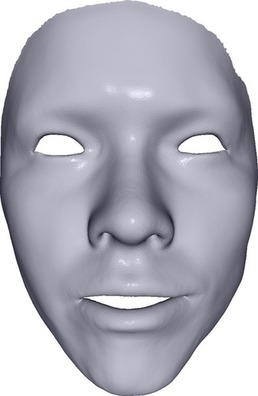}
\\
\includegraphics[height=.085\textheight]{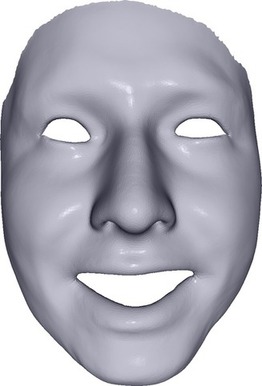} &
\includegraphics[height=.085\textheight]{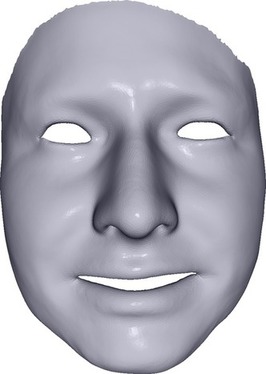} &
\includegraphics[height=.085\textheight]{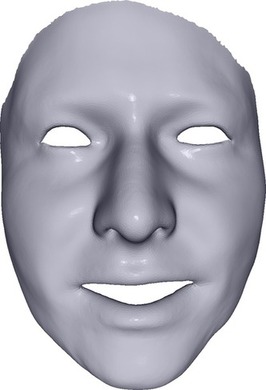} &
\includegraphics[height=.085\textheight]{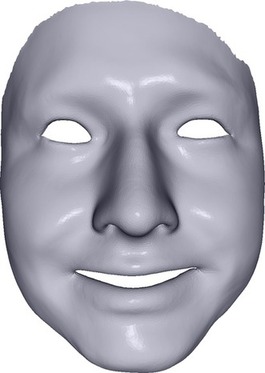} &
\includegraphics[height=.085\textheight]{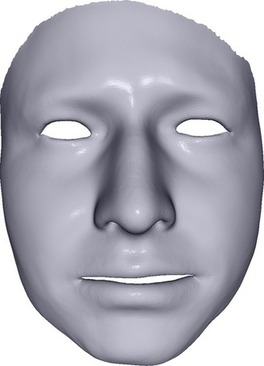} &
\includegraphics[height=.085\textheight]{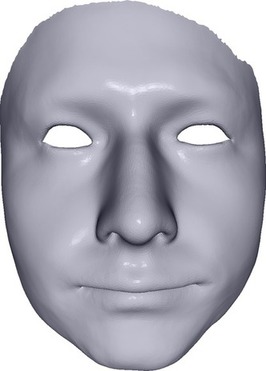} &
\includegraphics[height=.085\textheight]{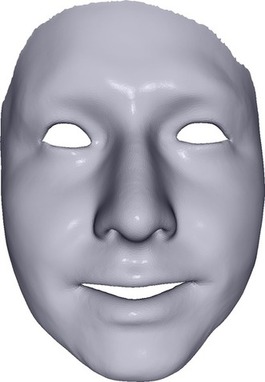} &
\includegraphics[height=.085\textheight]{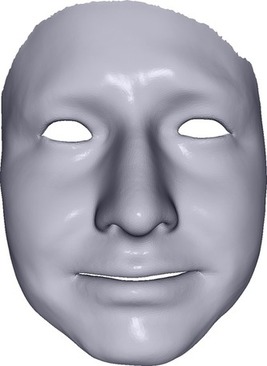}
\\
\includegraphics[height=.085\textheight]{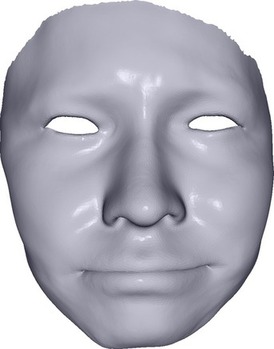} &
\includegraphics[height=.085\textheight]{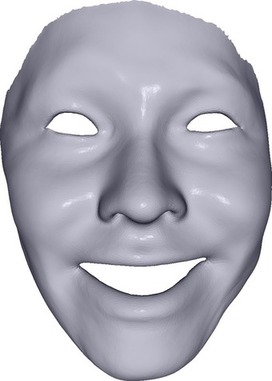} &
\includegraphics[height=.085\textheight]{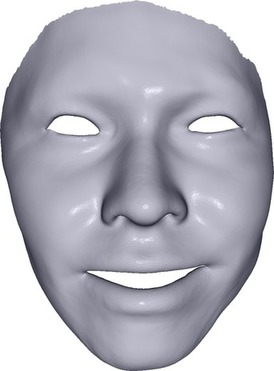} &
\includegraphics[height=.085\textheight]{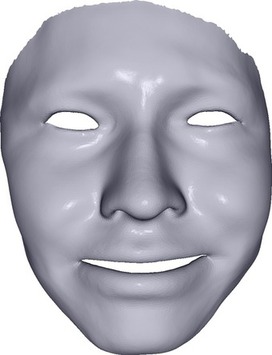} &
\includegraphics[height=.085\textheight]{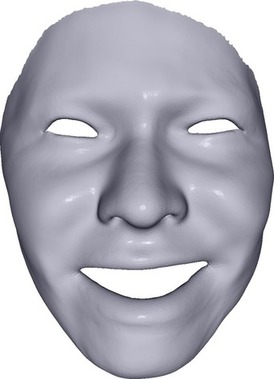} &
\includegraphics[height=.085\textheight]{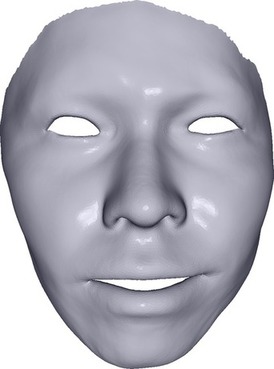} &
\includegraphics[height=.085\textheight]{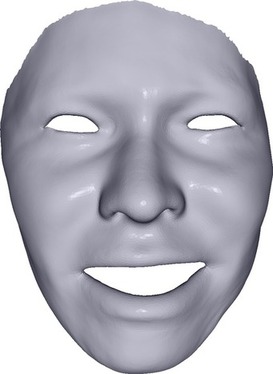} &
\includegraphics[height=.085\textheight]{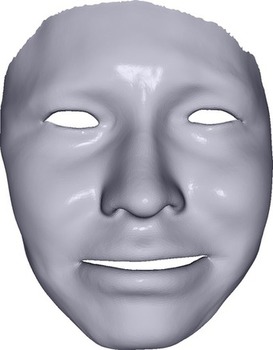}
\end{tabular}
\caption{Ours}
\end{subfigure}
\end{center}
\caption{Example of results used for identity decoupling and diversity evaluation, for the three compared methods. Each row shows samples with a same identity code, while the expression code is drawn randomly. Note the low variability in the generated samples for MAE, as also seen in Table 1.}
\label{fig:dec_id}
\end{figure}
\newcolumntype{B}{m{0.085\textwidth}}
\begin{figure}[h!]
\begin{center}

\begin{subfigure}[b]{\textwidth}
\centering
\begin{tabular}{BBBBBBBB}
\includegraphics[height=.085\textheight]{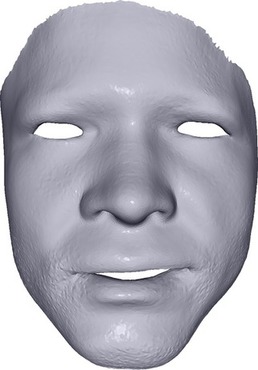} &
\includegraphics[height=.085\textheight]{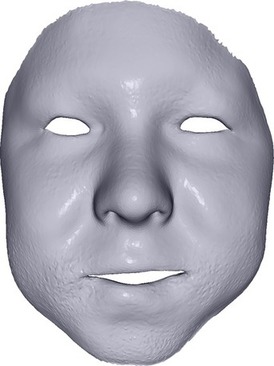} &
\includegraphics[height=.085\textheight]{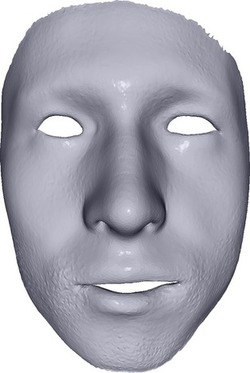} &
\includegraphics[height=.085\textheight]{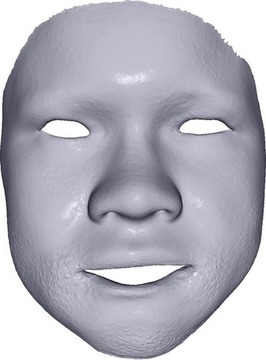} &
\includegraphics[height=.085\textheight]{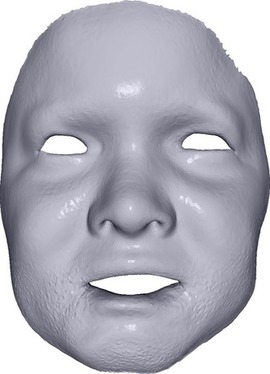} &
\includegraphics[height=.085\textheight]{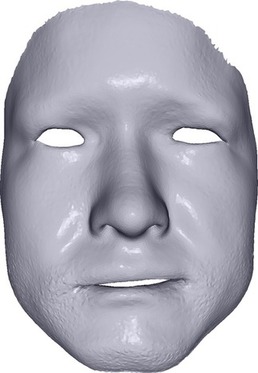} &
\includegraphics[height=.085\textheight]{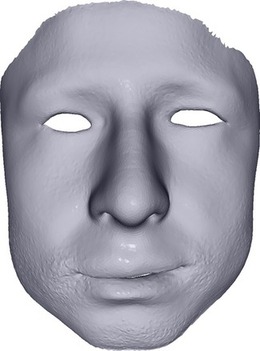} &
\includegraphics[height=.085\textheight]{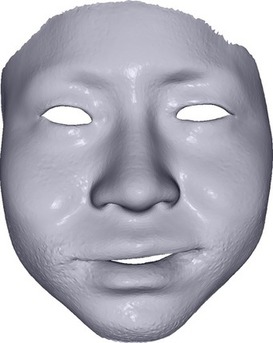}
\\
\includegraphics[height=.085\textheight]{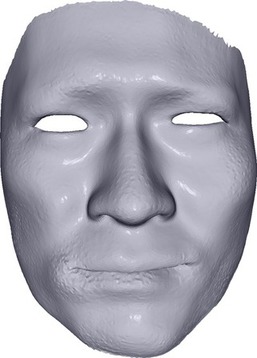} &
\includegraphics[height=.085\textheight]{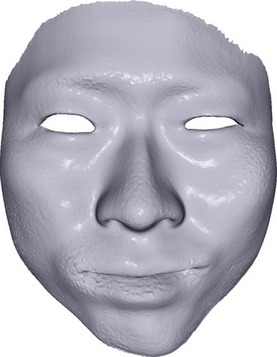} &
\includegraphics[height=.085\textheight]{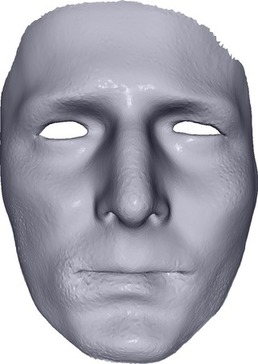} &
\includegraphics[height=.085\textheight]{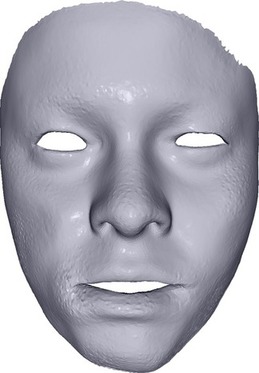} &
\includegraphics[height=.085\textheight]{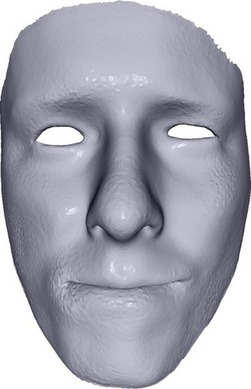} &
\includegraphics[height=.085\textheight]{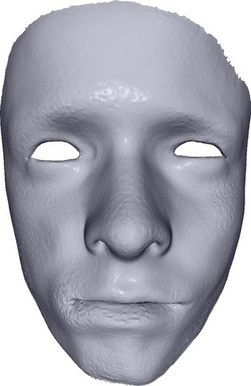} &
\includegraphics[height=.085\textheight]{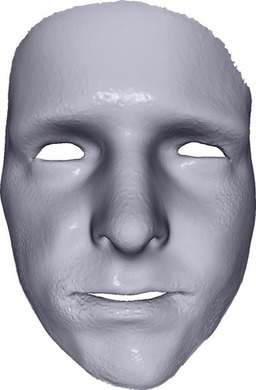} &
\includegraphics[height=.085\textheight]{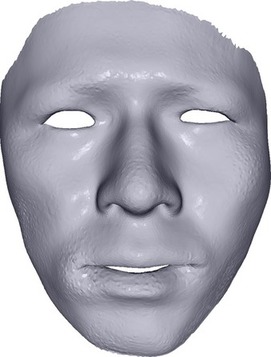}
\\
\includegraphics[height=.085\textheight]{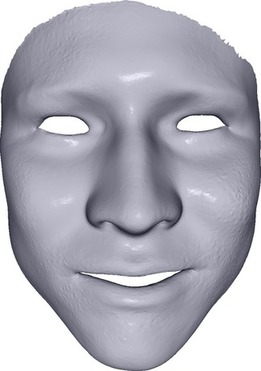} &
\includegraphics[height=.085\textheight]{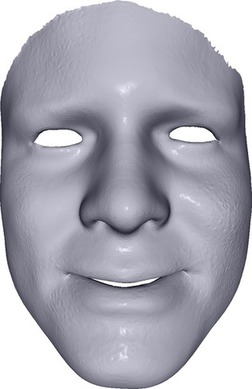} &
\includegraphics[height=.085\textheight]{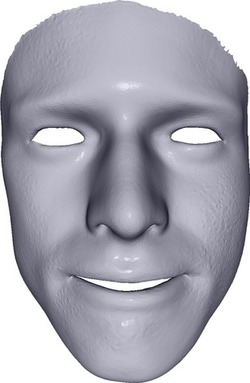} &
\includegraphics[height=.085\textheight]{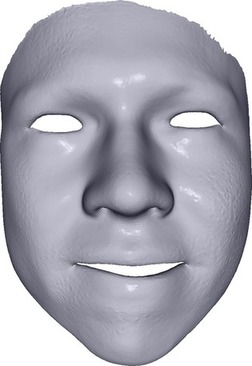} &
\includegraphics[height=.085\textheight]{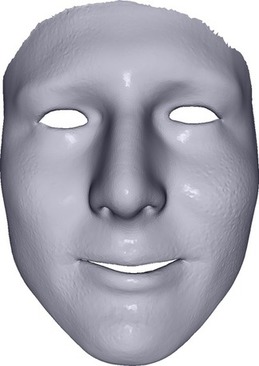} &
\includegraphics[height=.085\textheight]{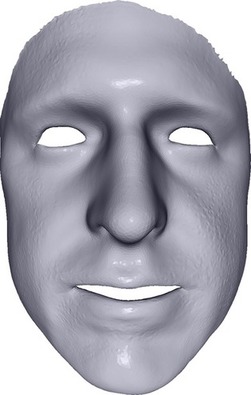} &
\includegraphics[height=.085\textheight]{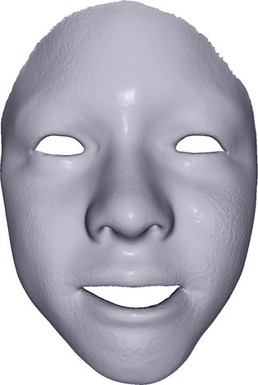} &
\includegraphics[height=.085\textheight]{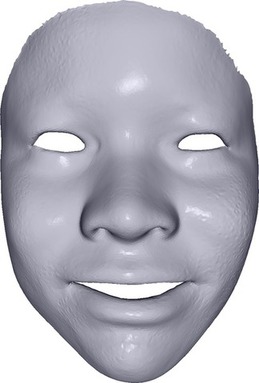}
\end{tabular}
\caption{COMA}
\end{subfigure}

\begin{subfigure}[b]{\textwidth}
\centering
\begin{tabular}{BBBBBBBB}
\includegraphics[height=.085\textheight]{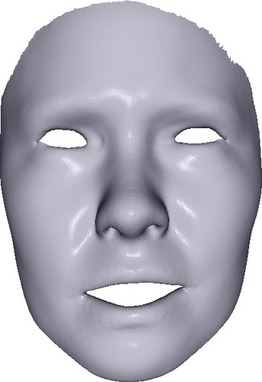} &
\includegraphics[height=.085\textheight]{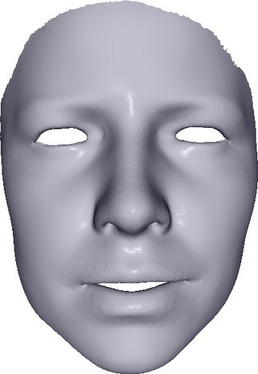} &
\includegraphics[height=.085\textheight]{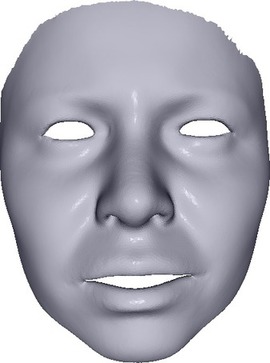} &
\includegraphics[height=.085\textheight]{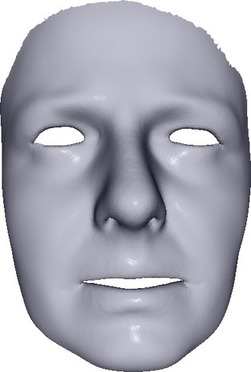} &
\includegraphics[height=.085\textheight]{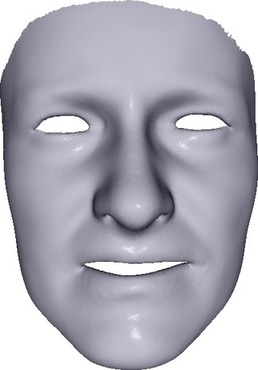} &
\includegraphics[height=.085\textheight]{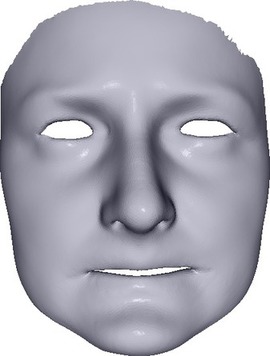} &
\includegraphics[height=.085\textheight]{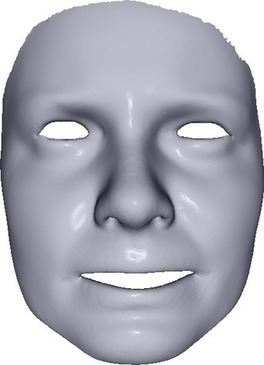} &
\includegraphics[height=.085\textheight]{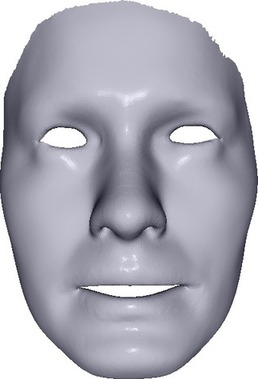}
\\
\includegraphics[height=.085\textheight]{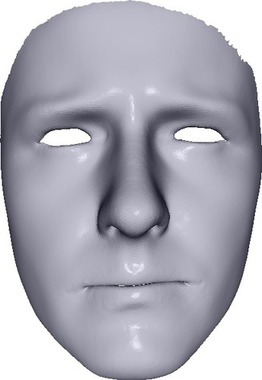} &
\includegraphics[height=.085\textheight]{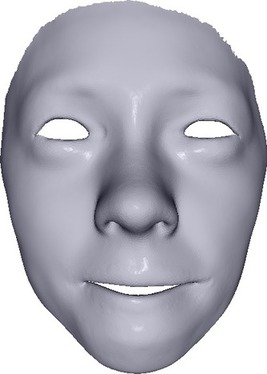} &
\includegraphics[height=.085\textheight]{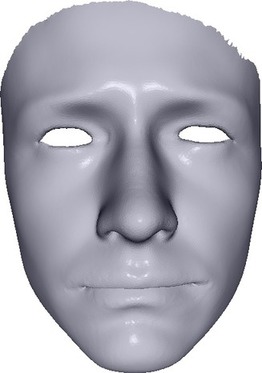} &
\includegraphics[height=.085\textheight]{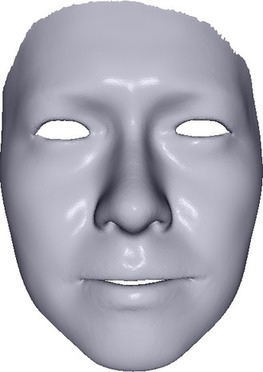} &
\includegraphics[height=.085\textheight]{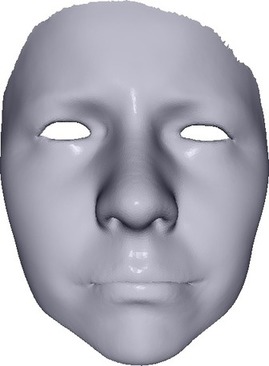} &
\includegraphics[height=.085\textheight]{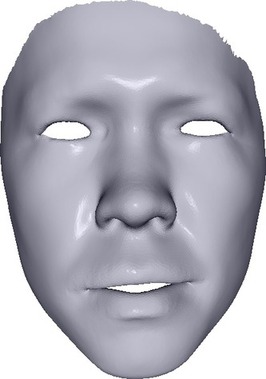} &
\includegraphics[height=.085\textheight]{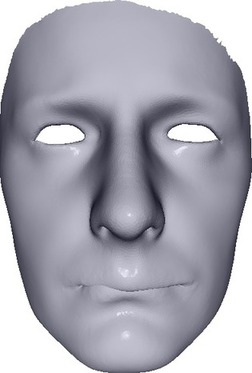} &
\includegraphics[height=.085\textheight]{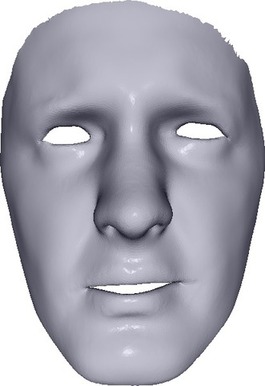}
\\
\includegraphics[height=.085\textheight]{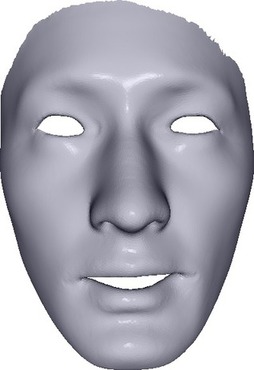} &
\includegraphics[height=.085\textheight]{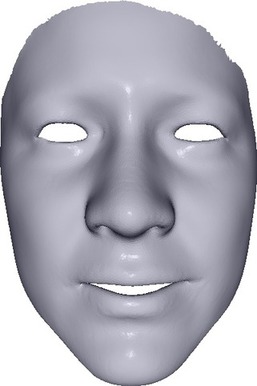} &
\includegraphics[height=.085\textheight]{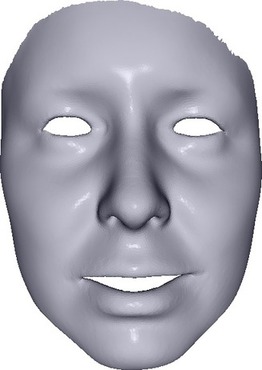} &
\includegraphics[height=.085\textheight]{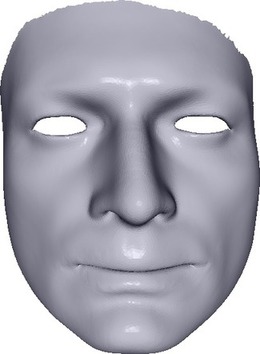} &
\includegraphics[height=.085\textheight]{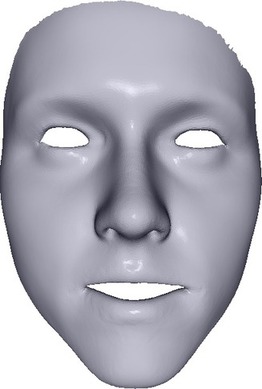} &
\includegraphics[height=.085\textheight]{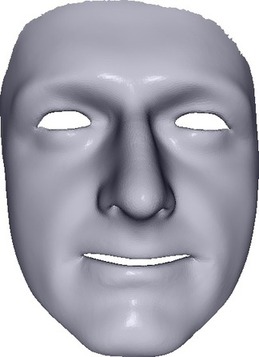} &
\includegraphics[height=.085\textheight]{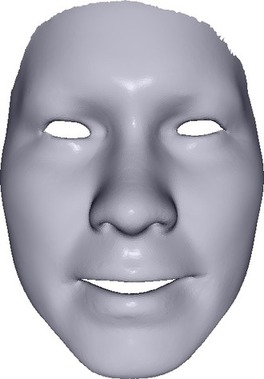} &
\includegraphics[height=.085\textheight]{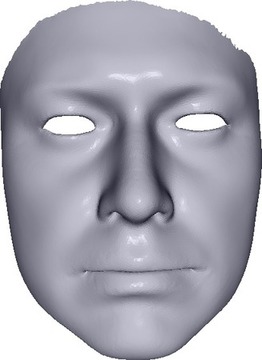}
\end{tabular}
\caption{MAE}
\end{subfigure}

\begin{subfigure}[b]{\textwidth}
\centering
\begin{tabular}{BBBBBBBB}
\includegraphics[height=.085\textheight]{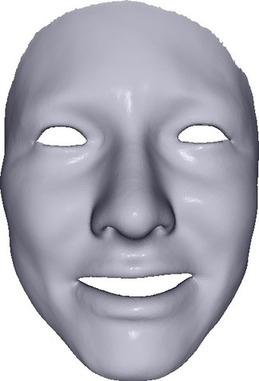} &
\includegraphics[height=.085\textheight]{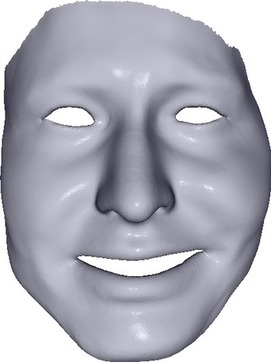} &
\includegraphics[height=.085\textheight]{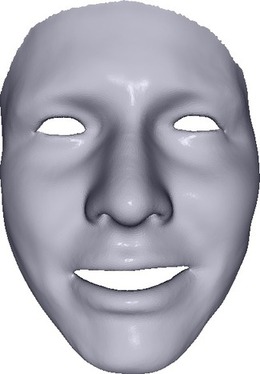} &
\includegraphics[height=.085\textheight]{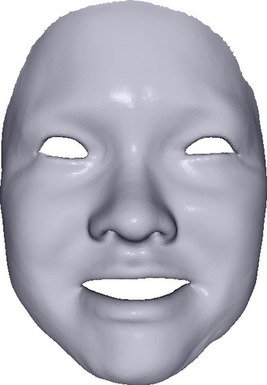} &
\includegraphics[height=.085\textheight]{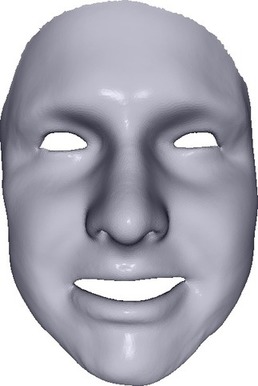} &
\includegraphics[height=.085\textheight]{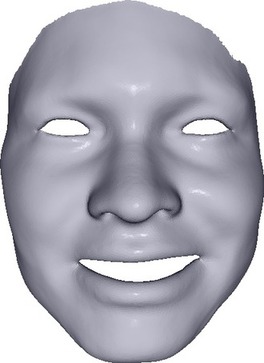} &
\includegraphics[height=.085\textheight]{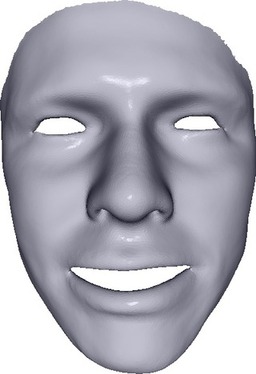} &
\includegraphics[height=.085\textheight]{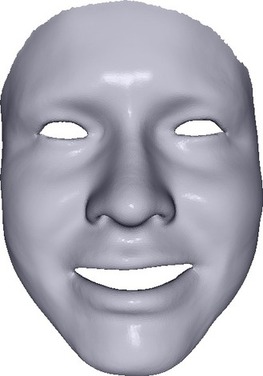}
\\
\includegraphics[height=.085\textheight]{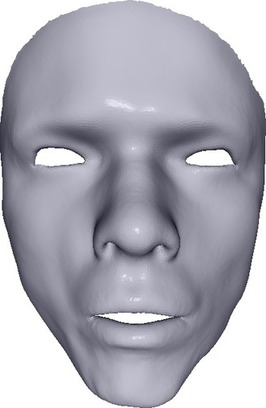} &
\includegraphics[height=.085\textheight]{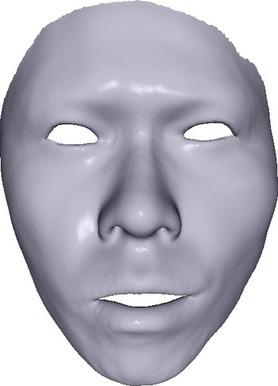} &
\includegraphics[height=.085\textheight]{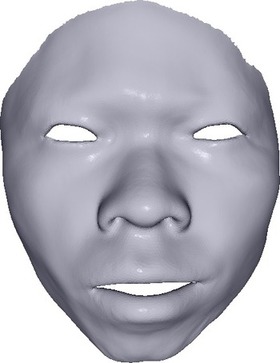} &
\includegraphics[height=.085\textheight]{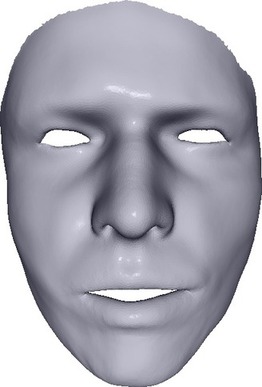} &
\includegraphics[height=.085\textheight]{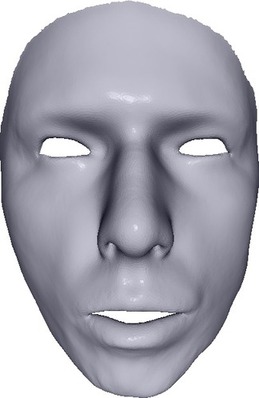} &
\includegraphics[height=.085\textheight]{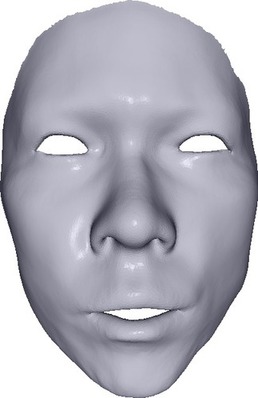} &
\includegraphics[height=.085\textheight]{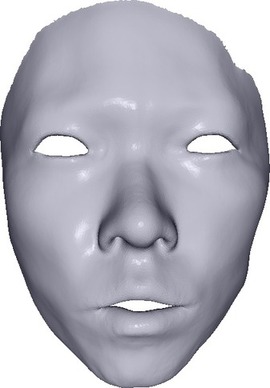} &
\includegraphics[height=.085\textheight]{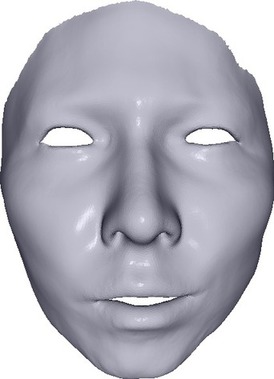}
\\
\includegraphics[height=.085\textheight]{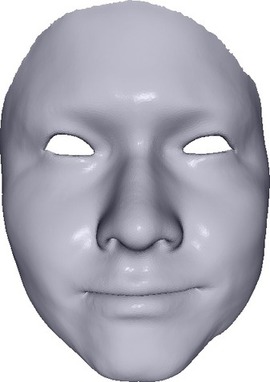} &
\includegraphics[height=.085\textheight]{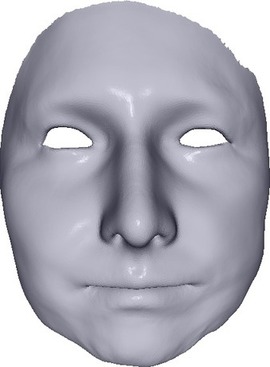} &
\includegraphics[height=.085\textheight]{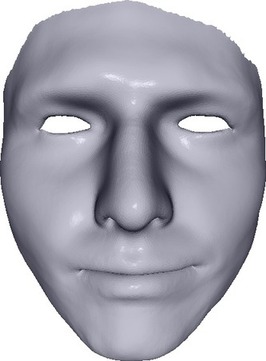} &
\includegraphics[height=.085\textheight]{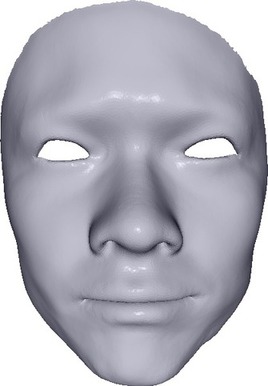} &
\includegraphics[height=.085\textheight]{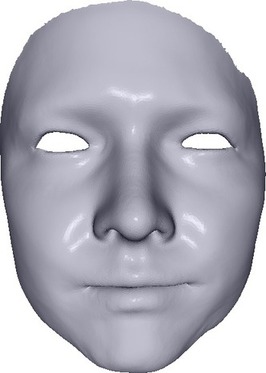} &
\includegraphics[height=.085\textheight]{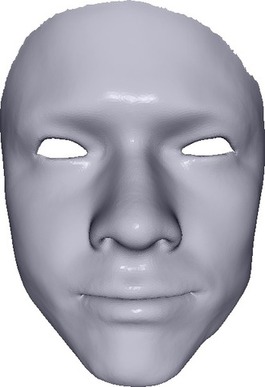} &
\includegraphics[height=.085\textheight]{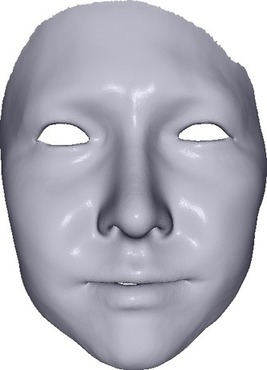} &
\includegraphics[height=.085\textheight]{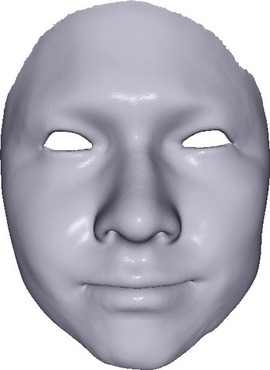}
\end{tabular}
\caption{Ours}
\end{subfigure}

\end{center}
\caption{Example of results used for expression decoupling and diversity evaluation, for the three compared methods. Each row shows samples with a same expression code, while the identity code is drawn randomly.}
\label{fig:dec_expr}
\end{figure}

\section*{Reconstruction of Sparse Data}

Figure~\ref{fig:reconstruction} shows qualitative results for the experiment in Table 2. The landmarks used for this evaluation are shown in Figure~\ref{fig:lmks}.
\begin{figure}[H]
\centering

\begin{subfigure}[b]{\textwidth}
\centering
\begin{tabular}{c c c c c}
Input & & With regularization & & No regularization \\ 
\\
\includegraphics[height=.11\textheight]{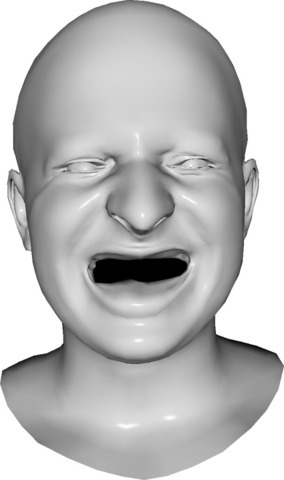} &
\qquad &
\includegraphics[height=.1\textheight]{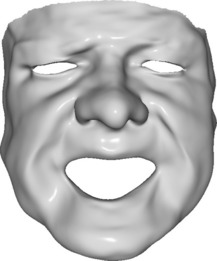}
\includegraphics[height=.1\textheight]{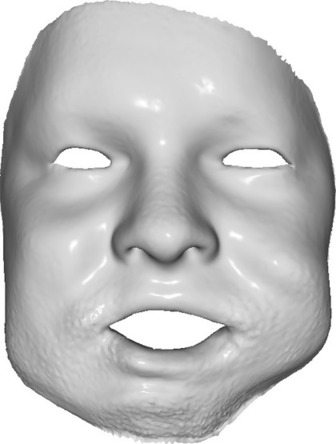}
\includegraphics[height=.1\textheight]{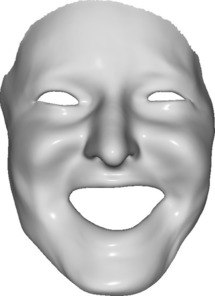} & 
\qquad & 
\includegraphics[height=.1\textheight]{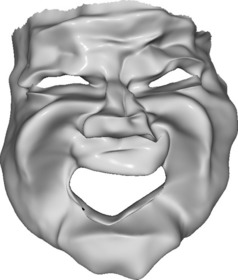}
\includegraphics[height=.1\textheight]{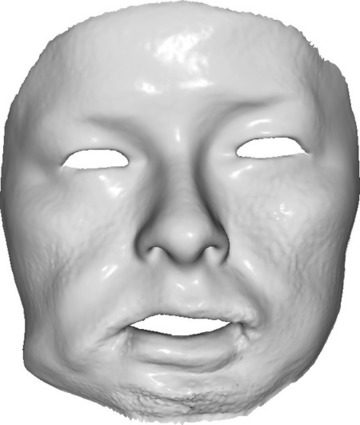}
\includegraphics[height=.1\textheight]{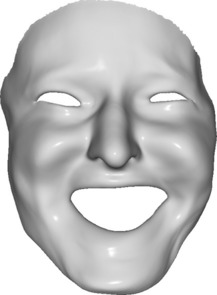}
\\
\includegraphics[height=.11\textheight]{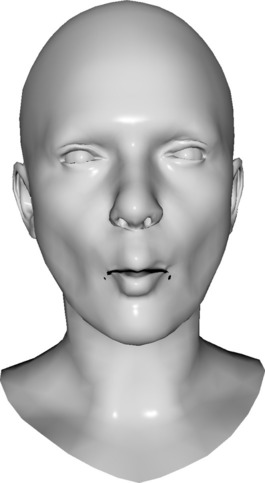} &
\qquad &
\includegraphics[height=.1\textheight]{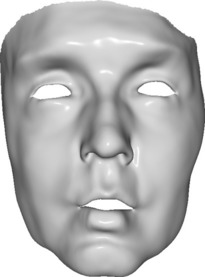}
\includegraphics[height=.1\textheight]{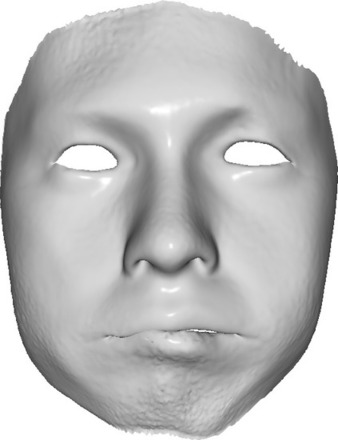}
\includegraphics[height=.1\textheight]{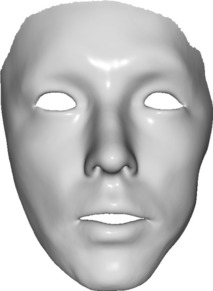} & 
\qquad & 
\includegraphics[height=.1\textheight]{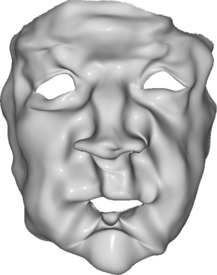}
\includegraphics[height=.1\textheight]{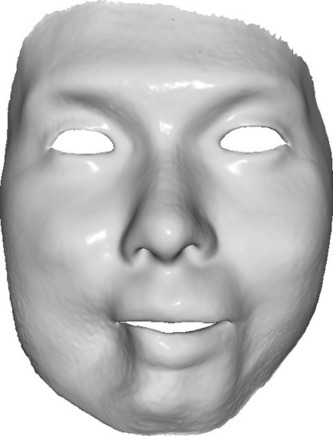}
\includegraphics[height=.1\textheight]{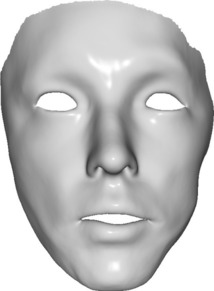}
\\
\includegraphics[height=.11\textheight]{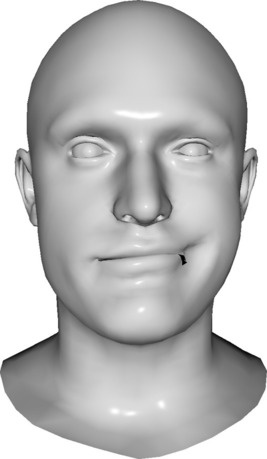} &
\qquad &
\includegraphics[height=.1\textheight]{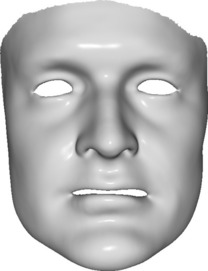}
\includegraphics[height=.1\textheight]{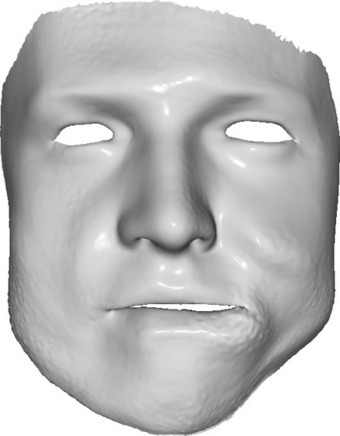}
\includegraphics[height=.1\textheight]{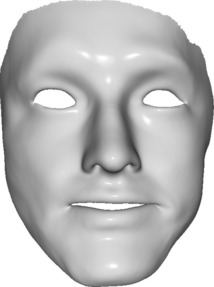} & 
\qquad & 
\includegraphics[height=.1\textheight]{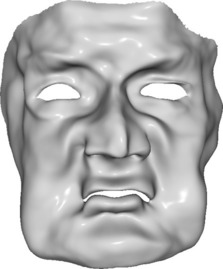}
\includegraphics[height=.1\textheight]{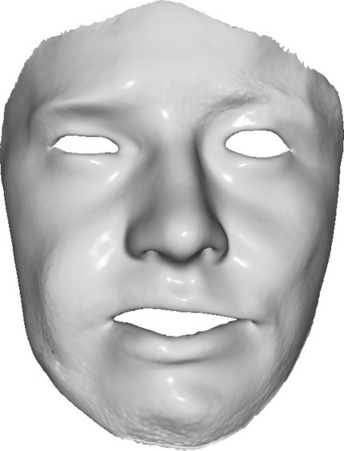}
\includegraphics[height=.1\textheight]{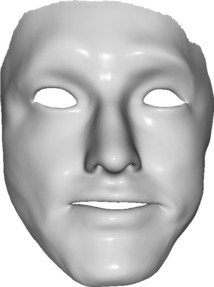}
\\
\includegraphics[height=.11\textheight]{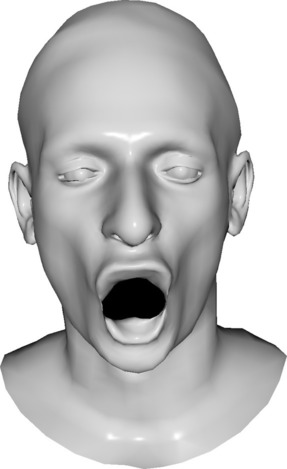} &
\qquad &
\includegraphics[height=.1\textheight]{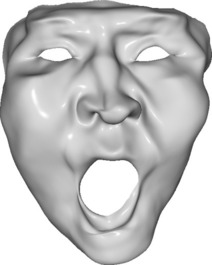}
\includegraphics[height=.1\textheight]{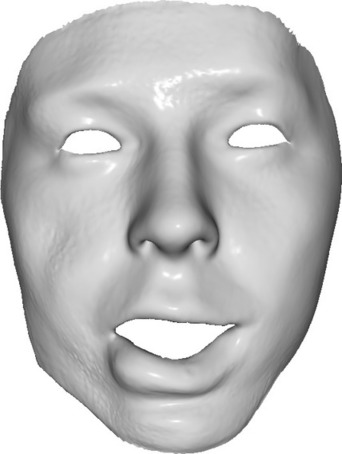}
\includegraphics[height=.1\textheight]{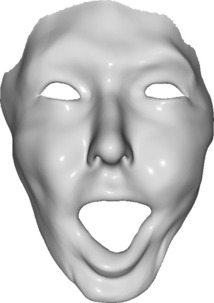} & 
\qquad & 
\includegraphics[height=.1\textheight]{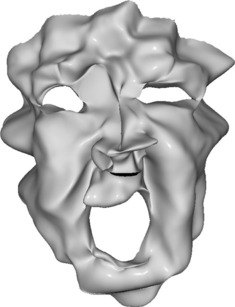}
\includegraphics[height=.1\textheight]{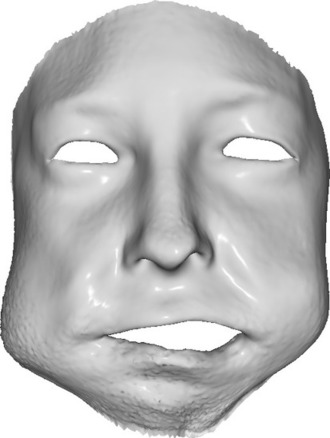}
\includegraphics[height=.1\textheight]{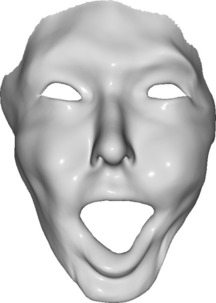}
\\
\end{tabular}
\caption{Comparison against MAE and COMA, with and without regularization. From left to right: MAE, COMA, our result.}
\label{fig:reconstruction}
\end{subfigure}

\begin{subfigure}[b]{\textwidth}
\vspace{20pt}
\centering
	\includegraphics[height=.13\textheight]{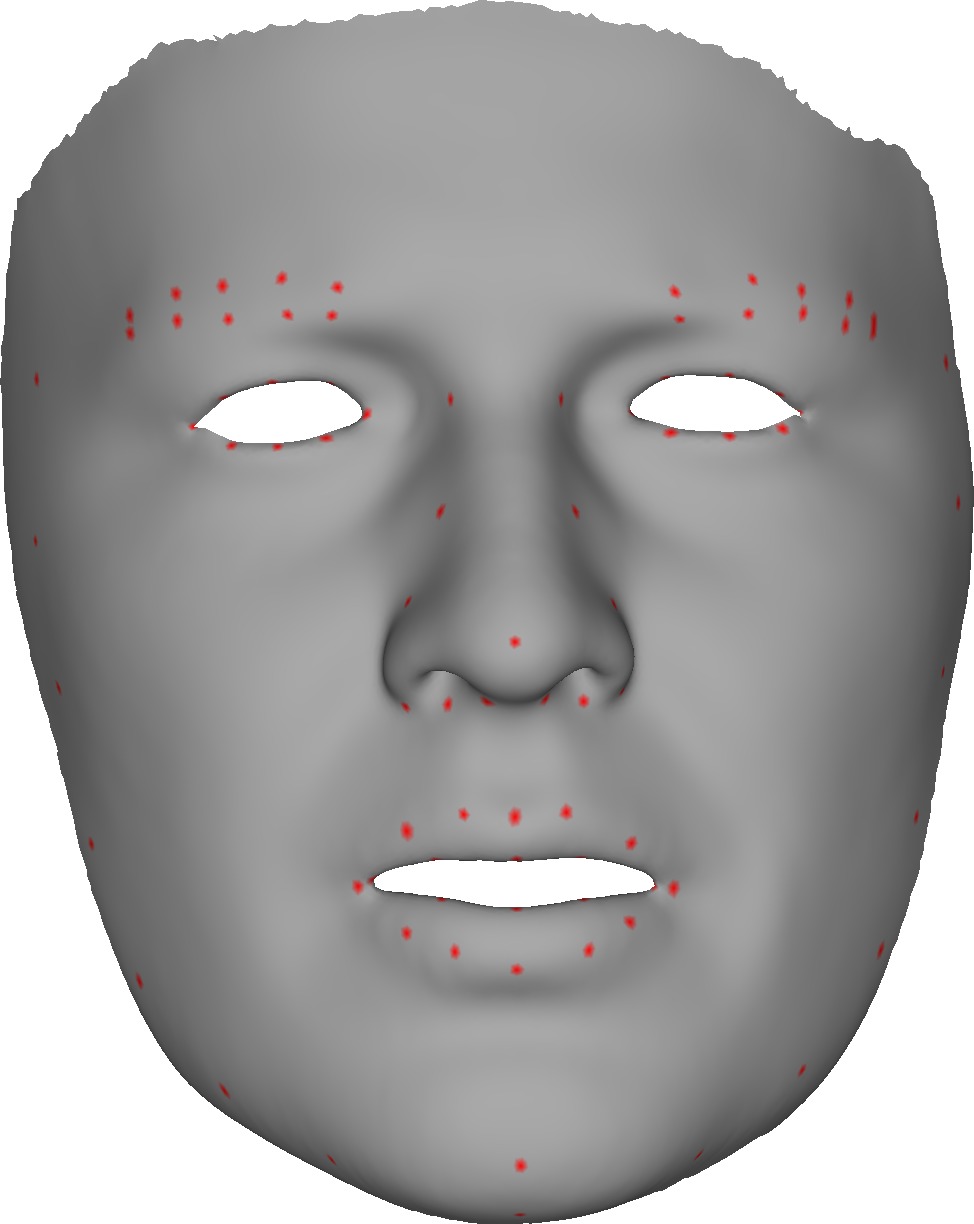}
\caption{$85$ landmarks used for fitting}
\label{fig:lmks}
\end{subfigure}

\caption{Reconstruction of sparse data}
\end{figure}


\newpage
\section*{Architecture Details}

In Figure~\ref{fig:arch} we show the architecture for the Generator and Discriminator used in this paper (the latter with the classification branches). Here, $d_{id}$, $d_{exp}$ and $d_{noise}$ are the dimensions for identity, expression and noise, respectively; $n_{id}$ is the number of distinct labels for identity, and $n_{exp}$ the number of distinct labels for expression. We use Leaky ReLU with a slope of $0.2$.

\begin{figure}[H]
\centering
\begin{subfigure}[b]{.8\textwidth}
\centering
\begin{tabular}{| l c r |}
\hline
Operation & Activation & Output Shape\\
\hline
$z \sim \mathcal{N}(0, I)$ & $-$ & $d_{id} + d_{exp} + d_{noise}$\\
\hline
Linear & LReLU & $512$\\
\hline
Linear & $-$ & $66387$\\
\hline
Reshape & $-$ & $22129 \times 3$\\
\hline
\end{tabular}
\caption{Generator}
\end{subfigure}

\begin{subfigure}[b]{.8\textwidth}
\vspace{15pt}
\centering
\begin{tabular}{| l c r|}
\hline
Operation & Activation & Output Shape\\
\hline
Input & $-$ & $22129 \times 3$\\
\hline
Geometry mapping & $-$ & $3 \times 64 \times 64$\\
\hline
\hline
\textbf{Common branch} & & \\
\hline
Conv $3 \times 3$ & LReLU & $16 \times 32 \times 32$\\
\hline
Conv $3 \times 3$ & LReLU & $32 \times 16 \times 16$\\
\hline
\hline
\textbf{Discriminator branch} & & \\
\hline
Conv $3 \times 3$ & LReLU & $64 \times 8 \times 8$\\
\hline
Conv $3 \times 3$ & LReLU & $128 \times 4 \times 4$\\
\hline
Reshape & $-$ & $2048$ \\
\hline
Linear & $-$ & $1$\\
\hline
\hline
\textbf{Identity branch} & & \\
\hline
Conv $3 \times 3$ & LReLU & $64 \times 8 \times 8$\\
\hline
Conv $3 \times 3$ & LReLU & $128 \times 4 \times 4$\\
\hline
Reshape & $-$ & $2048$ \\
\hline
Linear & $-$ & $n_{id}$\\
\hline
\hline
\textbf{Expression branch} & & \\
\hline
Conv $3 \times 3$ & LReLU & $64 \times 8 \times 8$\\
\hline
Conv $3 \times 3$ & LReLU & $128 \times 4 \times 4$\\
\hline
Reshape & $-$ & $2048$ \\
\hline
Linear & $-$ & $n_{exp}$\\
\hline
\end{tabular}
\caption{Discriminator and Classifiers.}
\end{subfigure}
\caption{Generator and Discriminator used for experiments in the paper}
\label{fig:arch}
\end{figure}

\section*{Decoupling Evaluation - Implementation Details}

We train the embedding networks using a Resnet-18 architecture with input images of size $224 \times 224$. The images contain the orthographic projection of the facial mesh, and the values in the RGB channels encode the normal direction of each vertex, as we found this to give better results than the UV images. The networks were trained using the datasets described in Section 5.2 with the provided labels. The threshold is selected such that it maximizes the accuracy on the validation set, while keeping the False Acceptance Rate (FAR) below $10\%$. We build the validation set by randomly choosing an equal number of positive and negative pairs from the testing split. We choose $0.14$ as threshold for identity, which achieves $98.66\%$ accuracy and a FAR of $1.21\%$. For expression we use $0.226$ as threshold, which achieves $84.2\%$ of accuracy and a FAR of $8.03\%$. 

%
%
%
%

\endgroup

\end{document}